\documentclass{article}
\usepackage{booktabs} 
\usepackage{multirow}
\usepackage[utf8]{inputenc}
\usepackage{authblk}
\usepackage{setspace}
\usepackage[margin=1.25in]{geometry}
\usepackage{graphicx}
\graphicspath{{./figures/}} 
\usepackage{subcaption}
\usepackage{amsmath}
\usepackage{xcolor}
\usepackage[authoryear]{natbib}
\usepackage{float}
\usepackage{hyperref} 

\title{Soybean Maturity Prediction using 2D Contour Plots from Drone based Time Series Imagery}

\author[1,$\dag$]{Bitgoeul Kim}
\author[2,$\dag$]{Samuel W. Blair}
\author[3]{Talukder Z. Jubery}
\author[3]{Soumik Sarkar}
\author[2]{Arti Singh}
\author[2*]{Asheesh K. Singh}
\author[1,3*]{Baskar Ganapathysubramanian}

\affil[1]{Department of Computer Engineering, Iowa State University, Ames, IA, USA.}
\affil[2]{Department of Agronomy, Iowa State University, Ames, IA, USA.}
\affil[3]{Department of Mechanical Engineering, Iowa State University, Ames, IA, USA.}
\affil[*]{Correspondence should be addressed to Baskar Ganapathysubramanian baskarg@iastate.edu and Asheesh K. Singh singhak@iastate.edu} 
\affil[$\dag$]{Bitgoeul Kim and Sam Blair contributed equally to this work.}

\date{}

\begin{document}

\maketitle

\begin{abstract}
Plant breeding programs require assessment and understanding of days to maturity for accurate selection and placement of entries in appropriate tests. Soybean breeding programs, in the early stages of the breeding pipeline, assign relative maturity ratings to experimental varieties that indicate their suitable maturity zones. Traditionally, the estimation of maturity rating value for breeding varieties has involved breeders manually inspecting fields and assessing maturity value visually. This approach relies heavily on expert judgment, making it subjective and demanding considerable time and effort. This study aimed to develop a machine-learning model for evaluating soybean maturity using UAV-based time-series imagery. Images were captured at three-day intervals, beginning as the earliest varieties started maturing and continuing until the last varieties fully matured. The data collected for this experiment consisted of 22,043 plots collected across three years (2021, 2022, and 2023) and represent relative maturity groups 1.6 - 3.9. We utilized contour plot images extracted from the time-series UAV RGB imagery as input for a neural network model. This contour plot approach encoded the temporal and spatial variation within each plot into a single image. A deep learning model was trained to utilize this contour plot to predict maturity ratings.
This model demonstrates a significant improvement in accuracy and robustness, achieving up to 85\% accuracy. We also evaluate the model's accuracy as the number of imaging time points is reduced, quantifying the trade-off between temporal resolution and maturity prediction. The predictive model offers a scalable, objective, and efficient means of assessing crop maturity, enabling phenomics and ML approaches to reduce the reliance on manual inspection and subjective assessment. This approach enables the automatic prediction of relative maturity ratings in a breeding program, saving time and resources. 
\end{abstract}


\section{Introduction}
Soybean (\textit{Glycine max} (L.) Merr.) is a major global crop with more than 133 million hectares grown annually, of which 37 million and 62 million are grown in North and South America, respectively \citep{fao_soybean_2023}. Soybean has a very wide global adaptation from the tropics to temperate regions and shows a large variation in days to maturity \citep{singh2021plant}. Soybean is a photoperiod-sensitive crop, and it is categorized into maturity groups (MGs), which are based on latitudes \citep{mourtzinis2017delineating}. These maturity groups are important for crop production, and are also of vital importance to plant breeders as they make strategic selection and placement of varieties in accordance with their appropriate maturity group. This allows soybeans to thrive in environments conducive to their full developmental cycle to maximize growth potential and seed yield in specific geographical regions. Incorrect maturity estimate complicates accurate selection in breeding programs. For example, if the breeding program inaccurately places an early maturity variety in the late MG test, it matures early and does not allow proper comparison with appropriate checks. It leads to a shorter growing period and pod shattering due to the significantly early maturity of the variety in a later maturity test that is harvested later, giving significantly lower yields. 

One method used by soybean breeders to determine soybean maturity is to collect visual relative maturity (RM) scores. This involves making manual assessments of soybean maturity relative to a check cultivar, which is typically commercially available to farmers and has a very well-established MG rating. The RM scores are then assigned to new soybean varieties with previously unknown or preliminary maturity scores based on the comparison to these check cultivars. This method requires breeders to make regular field visits, engage in meticulous note-taking, and perform visual assessments, which can vary subjectively from one rater to another. Therefore, traditional methods are very time-intensive and human resource demanding. This approach forms a critical bottleneck in the efficiency of soybean breeding operations, especially for large-scale programs with tens of thousands of plots that require maturity ratings. The necessity for multiple raters to monitor these plots exacerbates logistical and financial burdens, further complicating the maturity rating process.

The advent of uncrewed aerial systems (UAS) equipped with RGB and multispectral cameras, alongside the integration of machine learning (ML) algorithms, marks a new era in automating the RM rating process \citep{zhang2012application, guo2021uas}. These technologies present a viable solution to reduce the labor and time traditionally required for data collection by offering rapid, non-invasive means to assess crop maturity across vast agricultural landscapes \citep{singh2021plant}. For instance, in a study conducted in a large soybean breeding program, drones equipped with multi-spectral cameras were deployed to systematically capture the growth and color variation of soybeans across different developmental stages \citep{shammi2024application}. The images captured were then processed using ML algorithms to classify the maturity of the crops with a high degree of accuracy. This method streamlined the data collection process by covering hundreds of acres within a few hours and minimized human error associated with subjective visual assessments. 

As a result, the integration of UAS technology in soybean breeding programs has significantly accelerated the decision-making process for the selection and advancement of breeding varieties \citep{guo2021uas}. 
Studies have explored the various ways UAS technology can enhance soybean breeding programs, and optimize crop production \citep{herr2023unoccupied, singh2021high, sarkar2024cyber}. As the ML and deep learning (DL) models improve and become more sophisticated \citep{singh2016machine, singh2018deep}, further efficiencies and accuracy can be gained.

There has been a wide array of research using UAS remote sensing to develop high-throughput maturity rating systems for soybean breeding programs. The aim of these studies is largely to save time for breeding programs to make these ratings more quickly and with less human error. The sheer volume of work in this area demonstrates the value this type of system can add to a breeding program. Studies predicting soybean maturity have often relied on complex models, such as convolutional neural networks (CNNs) \citep{moeinizade2022applied, trevisan2020high, zhang2022monitoring}, while others have used simpler approaches like Random Forest classifiers \citep{yu2016development, yuan2019early}. Some research has also focused on using vegetative index thresholds to classify plots as either mature or immature \citep{wang2023mapping, narayanan2019improving}. Although many of these studies achieve high accuracy in maturity classification, they often work with relatively small datasets, typically fewer than 5000 plots \citep{borra2020closing, yu2016development, zhou2019estimation, wang2023mapping, zhang2022monitoring, volpato2021optimization, christenson2016predicting, hu2023uav}. Another common limitation in these works is their inability to classify soybean varieties into distinct maturity classes. Most approaches simply classify a plot as either mature or immature based on a single image and track the date when the plot reaches maturity \citep{zhou2019estimation, trevisan2020high, wang2023mapping, hu2023uav}. While this data can then be used to make classifications according to program definitions, our work integrates this step directly into the pipeline and offers a higher-resolution classification system with up to seven maturity classes. 

One experiment estimated soybean maturity using multi-spectral imagery captured at three specific time points within a single growing season, employing Partial Least Squares Regression to process a limited set of 130 features extracted from the images \citep{zhou2019estimation}. Similarly, others have used CNNs to analyze bi-weekly captured images, focusing on broad classifications or regression analyses of soybean maturity\citep{trevisan2020high, zhang2022monitoring}. These methods, however, typically utilized images only from one or two growing seasons and often relied on extracting key features rather than utilizing the entire image dataset \citep{zhou2019estimation, moeinizade2022applied}, potentially overlooking subtle yet critical environmental and developmental signals. 

Our research leverages a more comprehensive approach by utilizing the entire dataset, compressing all available image data from eight time series captured over three consecutive years from 2021 to 2023. This not only preserves the full spatial information but also allows for a more nuanced understanding of soybean maturity across varied environmental conditions. By introducing a sophisticated 7-class classification system, our study extends beyond the conventional four or five maturity stages commonly addressed in previous research, providing a more detailed and accurate predictive model of soybean maturity. This approach demonstrates the transformative potential of UAS in modern agriculture, not only in monitoring crop maturity but in significantly enhancing the precision and effectiveness of breeding programs.

Our study introduces a methodology for automating soybean RM ratings using UAS-derived RGB imagery and reports a new trait analysis approach derived from RGB UAS images. We extract hue values from time series images, constructing 2D contour plots that encapsulate time series, hue, and pixel count by hue as a novel phenotype input. 
This approach compressed the time series imagery of each plot into a single 2D contour plot, which is then used in the Convolutional Neural Network (CNN) dataset, allowing us to effectively reduce the complexity of high-dimensional data and providing a more concise yet informative analysis. Leveraging this input, our ML model classifies different RM groups, thereby developing an automatic system that marks an advancement by combining time-series imaging techniques with data-driven models. 

Additionally, this study utilizes UAS imagery to extract vegetative indices, specifically focusing on the Excess Greenness Index (ExG) \citep{woebbecke1995color}, to gain deeper insights into the relationship between soybean relative maturity (RM) and the rate of greenness loss. By using ExG, we can track the greenness of each soybean plot over time as it senesces. By examining the dynamics of greenness loss over time, we aim to uncover potential correlations between the rate of decline in ExG values and soybean yield. Understanding such relationships could provide an additional criterion for breeders to consider in their selection processes, potentially leading to the development of soybean varieties with optimized yield potentials. 

The objectives of this study are: (a) to develop a compact representation of UAS-based time-series imagery using contour plots; (b) to evaluate the performance of various ML models for predicting soybean maturity; (c) to identify the minimum number of drone flights required for accurate prediction, thereby optimizing cost-efficiency, and (d) to explore the relationship between the rate of greenness loss, as indicated by ExG decline, and yield outcomes, providing a new dimension to maturity assessment and selection criteria in soybean breeding.
Our methods elevate the precision and speed of maturity assessments and diminish the overall costs and labor demands associated with soybean breeding programs.

\section{Materials and Methods}
\subsection{Field Experiment and Data Collection}
The data collected for this experiment were collected from advanced soybean breeding yield trials near Boone, IA (42.020,-93.773, 339 meters above sea level). The dataset is comprised of imagery captured by UAS from six fields over a span of three years. This imagery includes data from two fields in 2021 representing the F$_5$ and F$_6$ filial generations (14,665 plots), two fields in 2022 representing the F$_6$ and F$_7$ generations (3,328 plots), and two fields in 2023 representing the F$_6$ and F$_7$ generations (4,050 plots), amounting to a total of 22,043 plots. Each plot consisted of two rows of planted soybeans with a row spacing of 76.2 centimeters and a seed spacing of 3.68 centimeters. The plot dimensions varied based on the generation: F$_5$ generation plots measured 2.13 meters in length with 0.91-meter alleys, while F$_6$ and F$_7$ generation plots measured 5.18 meters with 0.91-meter alleys. These trials were planted using a GPS-guided precision planter, ensuring accurate plot positioning and location tracking. The genetic material in these trials encompassed both elite and plant introduction (PI) varieties, with varieties spanning maturity groups (MG) from mid-MGI to late MGIII. 

Each year, soybean plots were seeded on a field that had bulk maize (\textit{Zea mays} L.) the previous year. Standard ground preparation methods were practiced. Each field was treated with a post-planting herbicide a month after planting. In addition to chemical weed control, manual weed control was done by routinely walking the plots to remove weeds. 

\begin{figure}[H]
    \centering
    \begin{subfigure}[b]{0.5\textwidth}  
        \centering
        \includegraphics[width=\textwidth]{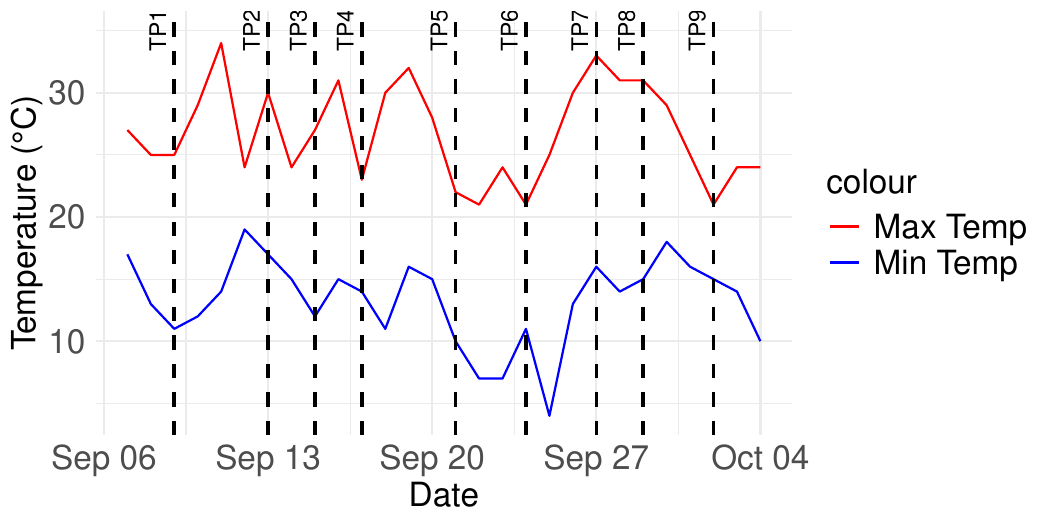}
        \caption{2021}
        \label{fig:weather_21}
    \end{subfigure}
    \vspace{0.5cm}  
    \begin{subfigure}[b]{0.5\textwidth}  
        \centering
        \includegraphics[width=\textwidth]{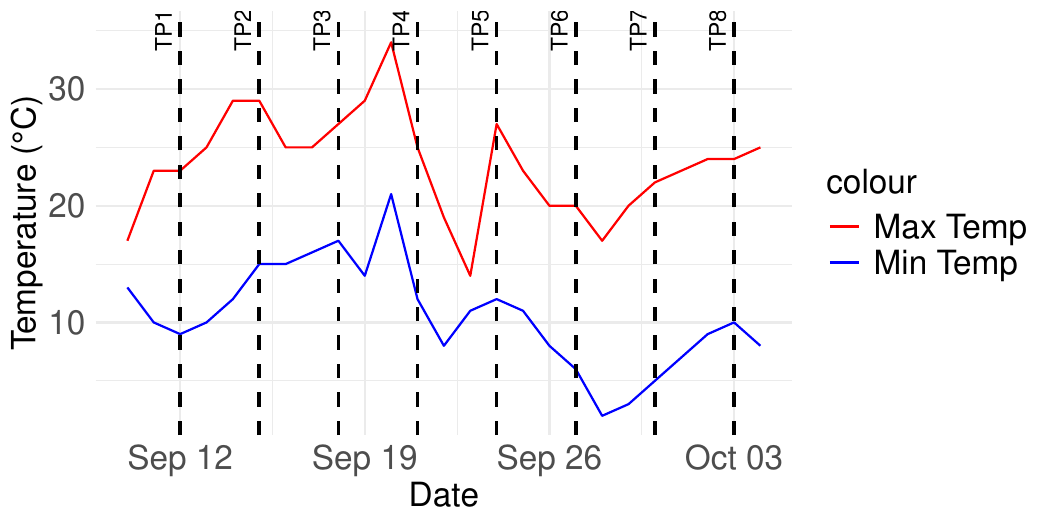}
        \caption{2022}
        \label{fig:weather_22}
    \end{subfigure}
    \vspace{0.5cm}  
    \begin{subfigure}[b]{0.5\textwidth}  
        \centering
        \includegraphics[width=\textwidth]{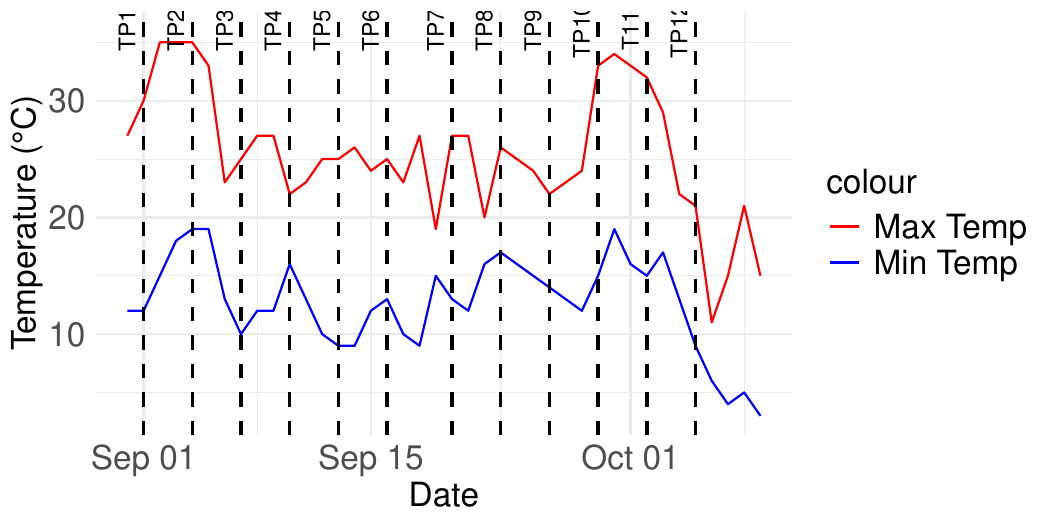}
        \caption{2023}
        \label{fig:weather_23}
    \end{subfigure}
    
    \caption{Maximum and minimum daily temperatures for 2021 \textbf{(a)}, 2022 \textbf{(b)}, and 2023 \textbf{(c)} during the period of time UAS flights were being conducted. Dashed lines represent UAS flight time points (TP).}
    \label{fig:weather_data}
\end{figure}

Ground truth relative maturity (RM) ratings were obtained through expert rater assessment. This involved planting check cultivars throughout each field, representing commercially available soybean cultivars with well-established maturity groups. Expert raters monitored these check cultivars on each visual data collection day to identify the cultivar that had most recently reached full maturity. The raters then assigned RM scores to new soybean varieties based on their phenological similarity to the check cultivars. Observations were conducted every two or three days to ensure that the RM of each new soybean variety, from earliest to latest, was accurately recorded. Yield data for these plots were obtained using either a Zurn plot combine (Zürn Harvesting, Schöntal-Westernhausen, Baden-Württemberg, Germany) or an Almaco plot combine (Almaco, Nevada, IA, USA) and have the units metric tons per hectare (MTH). Figure \ref{fig:weather_data} shows the daily max and minimum temperatures during the time frame UAS flights were being collected in 2021, 2022, and 2023.

\subsection{UAS Data Collection}
For imaging in 2021, a DJI Matrice 600 Pro (DJI, Shenzhen, China) equipped with a DJI Zenmuse X5 RGB camera and an Olympus M.Zuiko 45mm/1.8 lens (OM Digital Solutions, Shinjuku, Tokyo, Japan) was utilized. In 2022 and 2023, the imaging platform used was a DJI Inspire 2 with a DJI Zenmuse X5S RGB camera with an Olympus M.Zuiko 45mm/1.8 lens. Drone flights were taken every third day within 2 hours of solar noon, weather permitting. Flight plans were created using DroneDeploy (DroneDeploy, San Francisco, California, USA), with front overlap set to 70\% and side overlap to 80\%. The 2021 missions were flown at an altitude of 60 meters, achieving a ground sampling distance (GSD) of 0.5 cm/pixel. To maintain the same GSD in 2022 and 2023 with a higher resolution camera, the flight altitude was adjusted to 68 meters. Flight missions commenced before the earliest maturity check cultivar (MG of 1.9) reached full maturity. Each field was manually scouted to identify the onset of maturity in this early check cultivar, which marked the start of the imaging flight missions. Flights continued until the latest maturing varieties in the field reached full maturity. A retroactive start date was determined by retaining data from two flight dates prior to the recorded date when the early maturity check variety reached full maturity. A total of eight flights per field per year were retained. This ensured that the earliest RM group varieties up to the latest RM group varieties were recorded.

\subsection{Data Pre-Processing}
The UAS-captured images from each flight mission were processed and stitched into orthomosaics using Pix4Dmapper (Pix4D, Prilly, Switzerland). The orthomosaics were further processed using ArcGIS Pro (ESRI, Redlands, California, USA) and Python \citep{python}, where a grid was overlaid on each field's orthomosaic to segment each plot into distinct cells. Manual adjustments were made to the grid to ensure precise alignment of each plot within its respective cell. Once properly aligned, individual plot images were extracted in PNG format for each time point. As a result of this processing, a file was generated for each plot containing eight images, each associated with a corresponding ground truth RM rating.

\subsection{Greenness Slope Extraction}
For each plot obtained from the data pre-processing step, the Excess Greenness (ExG) index, an RGB color index, was computed. The ExG index is defined as:

\[
\text{ExG} = 2G - R - B
\]

where \(G\), \(R\), and \(B\) represent the green, red, and blue channels, respectively, in an RGB image. ExG values were calculated on a per-pixel basis across all plots for each time point and field, after which the mean ExG value for each plot at each time point was determined. ExG is extensively used in monitoring soybean plant growth \citep{borra2020closing}, and serves as a viable method for tracking soybean senescence. We chose this vegetative index specifically as a way to track plot greenness over time. It has also shown to be useful in differentiating grass and soil \citep{barbosa2019rgb} and estimating sugar cane yields \citep{khuimphukhieo2023use}. 

\begin{figure}[h]
    \centering
    \begin{subfigure}[b]{0.45\textwidth}
        \centering
        \includegraphics[width=\textwidth]{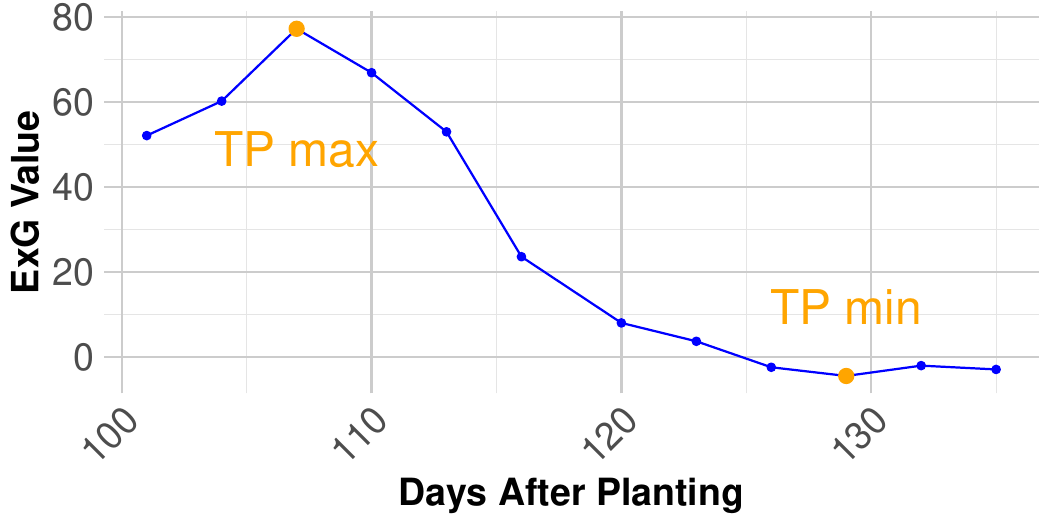}
        \caption{Plot with every timepoint captured by a UAS. TP max and TP min are highlighted in orange.}
        \label{fig:image1}
    \end{subfigure}
    \hfill
    \begin{subfigure}[b]{0.45\textwidth}
        \centering
        \includegraphics[width=\textwidth]{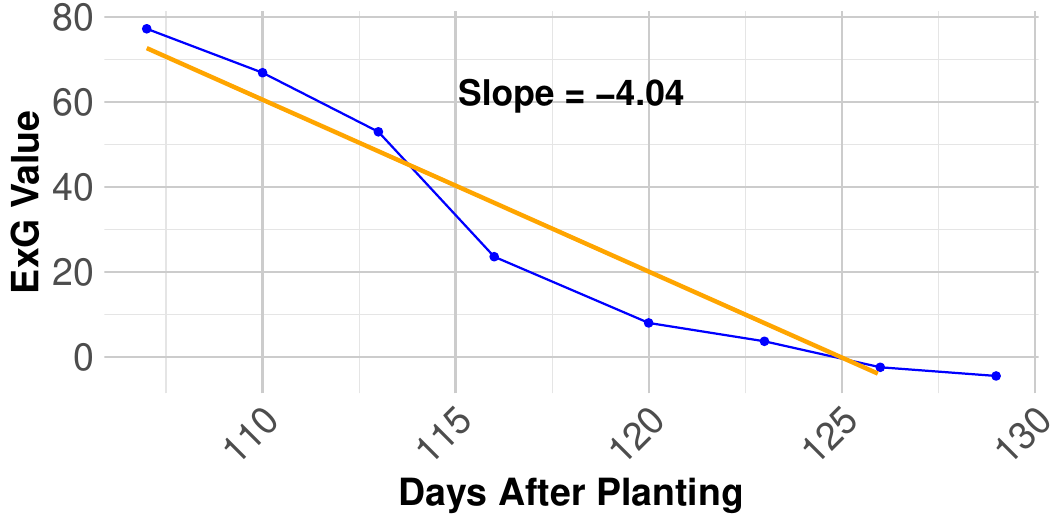}
        \caption{Subset of timepoints between and including TP max and TP min with regression line.}
        \label{fig:image2}
    \end{subfigure}
    \caption{Example of a random plot before (a) and after (b) subsetting. Points between and including the TP max and TP min are subsetted, and a linear regression is calculated for these points. The slope of this regression line is then calculated to represent the linear rate of greenness loss over time. }
    \label{fig:exg_slope}
\end{figure}

Subsequently, the time points corresponding to the maximum (TP max) and minimum (TP min) ExG values were identified for each plot. These two points were selected to approximate the start and end of the period during which greenness begins and ceases to decrease. Specifically, TP max represents the time point at which a soybean line is at maximum greenness (i.e., where it begins to senesce), while TP min represents the point at which the soybean line reaches its mature color. A subset of time points, spanning from TP max to TP min, was selected for further analysis. Linear regression was then applied to this subset to derive the slope, representing the linear rate of change in ExG value over time. This was done to observe differences in the rate of greenness loss between different RM classes. This slope was calculated for each individual plot belonging to the fields that represent the F$_6$ material from 2021, the F$_6$ and F$_7$ material from 2022, and the F$_6$ and F$_7$ material from 2023.  Figure \ref{fig:exg_slope} illustrates an example plot before and after processing. 

\subsection{Compact Representation of Time Series Data}

Using high-dimensional time-series image data obtained from soybean maturity assessment, we utilized an approach that gives a compact representation of time-series data in the context of soybean maturity. 

The following section outlines our methodology, including image preprocessing, feature extraction, and contour plot generation. These steps ensure that the high-dimensional dataset is both consistent and informative for the ML model.

\subsubsection*{Image processing}
We pre-process the original plot images by first cropping out unnecessary white edges. This cropping slightly changed the size of each plot. Next, all images were resized to consistent dimensions (here, 300 $\times$ 1000 pixels) using OpenCV library's \textit{resize} function. This ensures that the each plot has identical dimensions, allowing comparisons across plots in the dataset.

\subsubsection*{Extract feature}
The primary feature for assessing soybean maturity was the color of the crops, represented through the hue values in images. The color of a soybean plot typically transitions from green to yellow to brown as the soybeans mature, reflecting changes in the plant's physiology. This physiological change is reflected as variations in the hue value of the plot image, with decreasing hue values --from approximately 85 (green) down to 35 (yellow) and further down to 20 (brown) -- as the plot matures. To leverage this correspondence, we first convert transform the RGB format into an equivalent HSV image format (using OpenCV's in-built routines). The hue channel in the HSV format primarily represents the dominant wavelength (color) and is not influenced by changes in intensity or brightness. Since HSV separates chromaticity from luminance, variations in illumination mostly affect the Value (brightness) and Saturation (color intensity) channels, not Hue. This makes Hue relatively stable under varying lighting conditions. Therefore, Hue is a reliable measure for consistent color information in field imaging across different illumination scenarios. With the hue (H) channel readily available, we can now perform color analysis and tracking of temporal changes. We extracted the color distribution across each pixel in the plot images, allowing us to systematically calculate the number of pixels for each hue value. From this process, we obtained histograms per time-point per plot, illustrated in Figure \ref{fig:pipeline_workflow}. This time-series of histograms encodes the gradual change in hue, and thus serves as a good indicator of maturity rating.

\subsubsection*{Hue Contour Image as a Phenotype}
\begin{figure}
    \centering
    \includegraphics[width=1.0\linewidth]{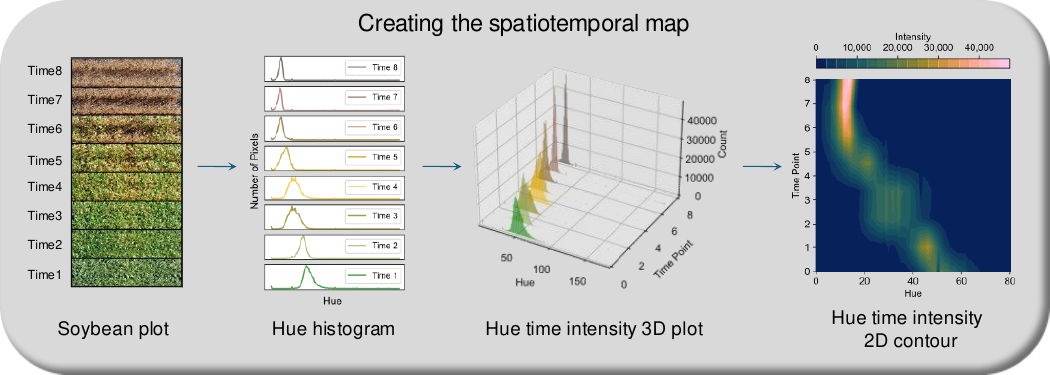}
    \caption{Overview of the workflow of 2D contour plot generation for the ML model. First, hue values and pixel counts are calculated for each time point. This workflow has feature extraction of the image as hue value, creating a 2D contour plot. For the final 2D contour plot, the hue range was cropped to 0-100. }
    \label{fig:pipeline_workflow}
\end{figure}

The histograms of hue at various time points were assembled into a single 3D figure (see Figure \ref{fig:pipeline_workflow}). This figure -- with hue value range (x-axis, 0 to 180), time (y-axis, represented by \( n \) time series) and the number of pixels corresponding to specific hue values at each time point for a plot (z-axis) -- encodes the spatio-temporal progression of the soybean plot. This 3D plot is converted into an image to leverage the ability to use sophisticated image based deep learning architectures. This conversion involved replacing the z-axis, which represented pixel counts, with a color map. The pixel count data were thus encoded as different colors in the 2D contour plot, allowing for a clearer and more interpretable representation of the data across different time points. We use the 'Batlow' colormap \citep{crameri2020misuse} for creating the 2D contour plots. Such perceptually uniform colormaps ensure minimum data distortion and visual errors, while ensuring clarity. The hue value range usually spans from 0 to 180, but our dataset showed no values beyond 100. We, therefore, crop the contour plot images at this hue threshold to eliminate irrelevant data areas, focusing analysis and visualization on the significant range.

Figure \ref{fig:pipeline_workflow}(right) shows a representative hue contour image for a specific soybean plot. This image encodes how the hue (intensity variations from left to right) changes with time (intensity variations from top to bottom), with the peak intensity moving from green to yellow to brown. 
We also created these hue contour images using varying number of time-series data. This allows us to investigate the tradeoff between number of UAV collection time-points and maturity prediction. Specifically, we created different contour plots by varying the number of time series data subsets: 8, 6, 4, and 3 time points. 

Through the processes described above, in 2023, we successfully generated a total of 3,194 2D contour plot images; in 2022, we generated 3,327 images; and in 2021, we generated 7,974 images. These images serve as the input dataset for our model. Plots that had missing rows due to planting error or did not emerge were not utilized in this dataset. This comprehensive collection and selection process ensured that only the most relevant and unbiased data was used, thereby enhancing the overall accuracy and reliability of our predictive analytics.

\subsection{Assigning Class Labels to Plots} 

Each of the plots is human rated with a maturity rating, that ranges from 1.6 to 3.9. These ratings are binned into discrete classes labels. We evaluated various binning strategies (Table \ref{tab:maturity_stages})
to investigate multiple scenarios due to the breeding program's interest in classifying more coarse to finer classification categories.
Thus, we consider binning the maturity ratings into scenarios with 4, 5, and 7 classes. For example, according to the criteria outlined in Table \ref{tab:maturity_stages}, a plot with a maturity value of 1.9 would be assigned to class 1 when categorized into five classes.  
We next train classifiers for each of these multi-class classification problems.

\begin{table}[ht]
\centering
\caption{Various Class Schemes Across Soybean Maturity Stages}
\label{tab:maturity_stages}
\begin{tabular}{@{}ccccccccc@{}} 
\toprule
& \multicolumn{8}{c}{Classification} \\
\cmidrule{2-9}
Maturity Rating Value & 1.6-2.0 & 2.1-2.3 & 2.4-2.6 & 2.7-2.9 & 3.0-3.2 & 3.3-3.5 & 3.6-3.7 & 3.8-3.9 \\
\midrule
7-Class & 1 & 2 & 3 & 4 & 5 & 6 & 6 & 7 \\
5-Class & 1 & 2 & 2 & 3 & 4 & 4 & 4 & 5 \\
4-Class (1st) & 1 & 2 & 2 & 3 & 4 & 4 & 4 & 4 \\
4-Class (2nd) & 1 & 2 & 2 & 3 & 3 & 4 & 4 & 4 \\
\bottomrule
\end{tabular}
\end{table}

\subsection{Model Development and ML pipeline}

\subsubsection*{Dataset Balancing and Augmentation}
\begin{figure}[H]
    \centering
    \begin{subfigure}[b]{0.49\linewidth}
        \centering
        \includegraphics[width=\linewidth]{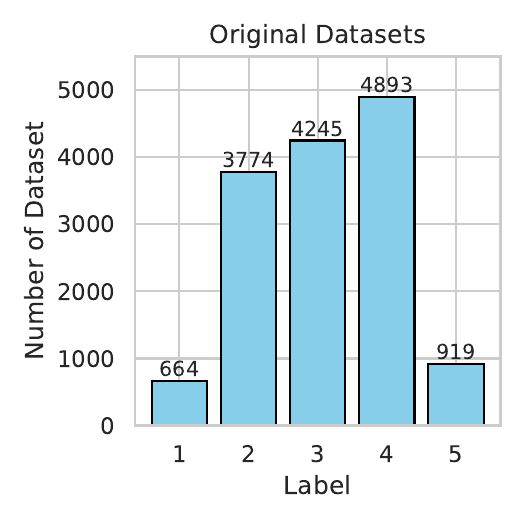}
        \caption{Original Dataset Before SMOTE}
        \label{fig:original_imbalanced}
    \end{subfigure}
    \hfill
    \begin{subfigure}[b]{0.49\linewidth}
        \centering
        \includegraphics[width=\linewidth]{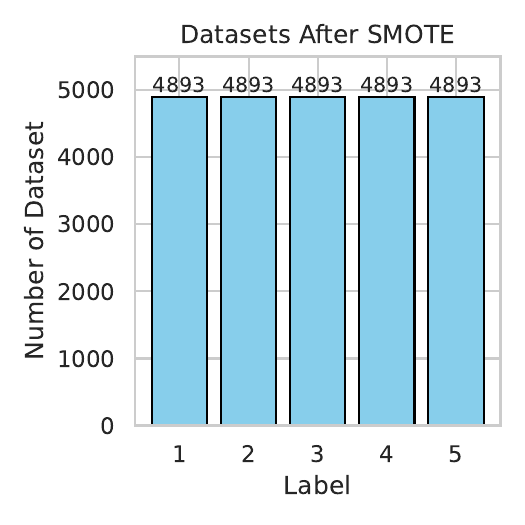}
        \caption{Resampled Dataset After SMOTE}
        \label{fig:aftersmote}
    \end{subfigure}
    \caption{(a) Original Dataset Before SMOTE: This graph illustrates the class imbalance present in the original dataset (for the 5 class prediction scenario). The y-axis represents the number of samples in each class, while the x-axis labels each class from 1 to 5. Class 1 has the fewest samples (664), while Class 4 has the most (4893). (b) Resampled Dataset After SMOTE: This graph shows the dataset after applying SMOTE to balance the number of samples across all classes. Each class now has an equal number of samples (4893), as indicated on the y-axis, with class labels shown on the x-axis from 1 to 5.}
    \label{fig:imbalance_balance_dataset}
\end{figure}

We observe significant class imbalance in the data. To address this, we first split our dataset into training, validation, and test sets with a ratio of 80\%, 10\%, and 10\%, respectively. We then tackled the class imbalance by applying standard synthetic data generation techniques and various data augmentation strategies exclusively to the training dataset.
To effectively address the challenge of class imbalance, which is known to adversely affect model performance, we employed the Synthetic Minority Over-sampling Technique (SMOTE). SMOTE is a widely used technique that generates synthetic instances by interpolating between existing minority class instances, as illustrated in Figure \ref{fig:imbalance_balance_dataset}. This process helps achieve balanced class distribution, enhancing model training and improving accuracy \citep{fernandez2018smote}.

For example, a dataset with five classes, as shown in Figure \ref{fig:original_imbalanced}, has a significant lack of data, especially in labels 1 and 5. In this case, the SMOTE technique was applied to create synthetic data for the classes with a smaller data size. As a result, the data increased evenly across all classes shown in Figure \ref{fig:aftersmote}, which contributed to improving the learning efficiency and overall performance of the model.

\begin{figure}[htbp]
  \centering
  \begin{subfigure}[b]{0.30\textwidth}
    \includegraphics[width=\textwidth]{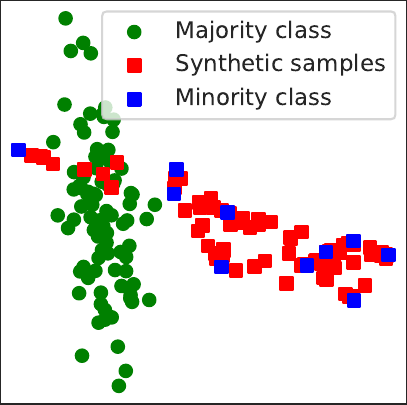}
    \caption*{(a)}
    \label{fig:image1}
  \end{subfigure}
  \hfill
  \begin{subfigure}[b]{0.65\textwidth}
    \centering
    \includegraphics[width=0.46\textwidth]{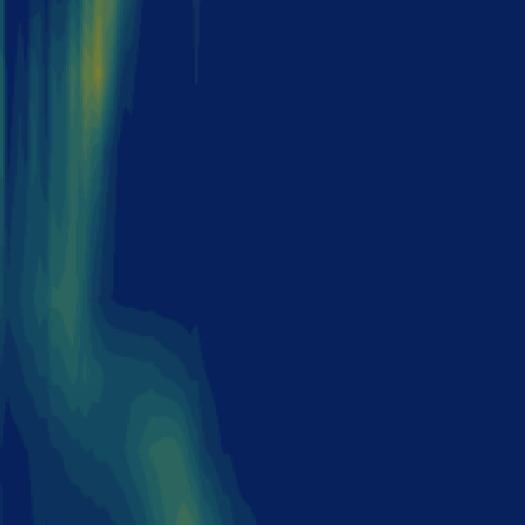}
    \hfill
    \includegraphics[width=0.46\textwidth]{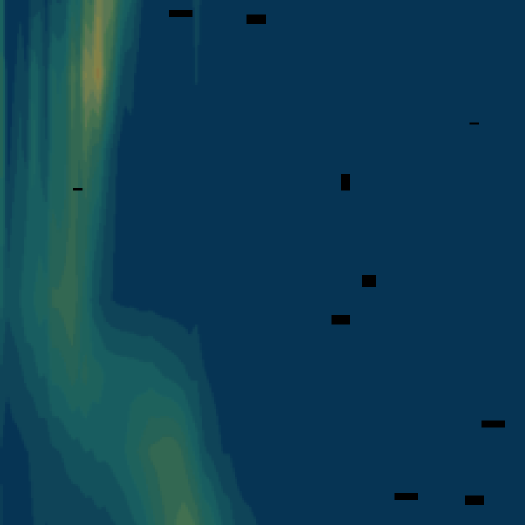}
    \caption*{(b)}
    \label{fig:subimages}
  \end{subfigure}
  \caption{(a) This figure illustrates an example of how SMOTE works with two classes. It shows the distribution of the majority class (green), minority class (blue), and the synthetic samples generated by SMOTE (red) to balance the class distribution.  (b) The left image is the original contour plot image. On the right side, there is the augmented image, where random masking and color-jitter are applied with the parameters: brightness = 0, contrast = 0.1, saturation = 0.2, and hue = 0.1. This augmentation introduces subtle variations in color and masks certain regions of the image, enhancing dataset diversity.}
  \label{fig:three_images}
\end{figure}

Furthermore, data augmentation techniques are specifically chosen based on the characteristics of our image data, which has informative values along both the x and y axes. The x-axis represents the hue range, while the y-axis indicates the sequential order of time series data. Due to the structured nature of our dataset, traditional augmentation methods such as cropping and rotating could potentially distort the meaningful attributes along these axes, leading to misleading training data and poor model performance. Therefore, we opted for color-jittering and random masking, which are less invasive but equally effective techniques for enhancing model generalizability without compromising the integrity of the data, as illustrated in Figure \ref{fig:three_images}(b). Color-jittering modifies the brightness, contrast, saturation, and hue of images randomly, thereby preparing the model to handle variations in lighting conditions and color distributions that may occur in real-world scenarios. Random masking, conversely, involves obscuring parts of the images randomly. This method simulates occlusions and varying degrees of visibility that may be encountered in practical applications, training the model to also focus on less obvious features and thus enhancing its ability to generalize from incomplete or partially visible data.
By employing these augmentation techniques on the training dataset, as illustrated in Figure \ref{fig:three_images}(b), we ensure that the model is not only trained on a broader spectrum of conditions but also develops robustness to variations in input data. This approach has been demonstrated to improve predictive performance across diverse scenarios, which is critical for applications in fields where data may not always be perfect or complete \citep{shorten2019survey}.

\subsubsection*{Model Training}
We experimented with various CNN architectures such as ResNet, VGG, and MobileNet, and focused primarily on ResNet34 due to its superior performance. We trained our model without using pre-trained weights from datasets like ImageNet. This approach was based on the significant differences between our contour images and the typical natural images in ImageNet, which could limit the effectiveness of pre-trained weights. Training from scratch allowed our model to learn features specific to our dataset, leading to better adaptation and performance in predicting soybean maturity.
We employed the Cross-Entropy Loss function, which is particularly suitable for classification tasks \citep{demirkaya2020exploring}. This loss function measures the performance of the model by comparing the predicted probability distribution of the classes to the actual distribution, providing a measure of how well the model's predictions align with the true labels. The Cross-Entropy Loss function, also known as log loss, is defined as follows:

The Cross-Entropy Loss function, also known as log loss, is defined as follows:

\[
\text{Loss} = -\sum_{i=1}^{N} y_i \log(\hat{y}_i)
\]

where \( y_i \) represents the true label for the \(i\)-th sample, \( \hat{y}_i \) represents the predicted probability for the \(i\)-th class, and \( N \) is the number of classes. This loss function penalizes incorrect classifications more severely when the predicted probability is far from the actual label, thereby encouraging the model to produce probabilities that are as close as possible to the true labels.

Our model training was conducted using PyTorch on an NVIDIA A100-SXM4-80GB GPU. The A100 GPU offers 80 GB of memory and is built on the NVIDIA Ampere architecture. The training environment included CUDA version 11.8 and driver version 520.61.05. 
The models were trained using a learning rate of 1e-4 and a batch size of 32. The training was conducted over 200 epochs to ensure sufficient learning and adaptation to the complex patterns in our data. 

\subsubsection*{Modeling with Hierarchical and Multi-Temporal Data}
We implemented a hierarchical classification method to predict the maturity stages into a structured order. This method not only simplifies the initial classification process by grouping stages into broader categories but also enhances the precision in distinguishing among closely related stages. Next, we exploited multi-temporal data by experimenting with different subsets of time series images, aiming to evaluate the model's performance under various data availabilities.

\paragraph{Hierarchical Classification Approach :}
Our hierarchical classification framework was inspired by previous studies that utilized a hierarchical structure to manage complexity in class distinctions \citep{seo2019hierarchical,naik2017real}. For the case of 7-class classification, we categorized the seven maturity stages into four main groups: (1), (2, 3), (4, 5), and (6, 7). Initially, the model classified samples into these broader groups corresponding to high-level maturity categories. Subsequent classification within these groups then focused on distinguishing between the finer stages. For example, within the (2, 3) group, the model determined whether a sample was at stage 2 or 3. This two-tiered approach significantly enhanced the accuracy and interpretability of our classifications, reducing the overall complexity faced at each classification level.

\paragraph{Utilization of Multi-Temporal Data :}
We next explored the impact of varying time series data on model performance. By selecting subsets containing 6, 4, and 3 images from the original set of eight time series data points, we assessed how the number of temporal observations affects the model's ability to classify maturity stages accurately. 

First, we selected images distributed across the entire time series:
    \begin{itemize}
        \item 6 images : 1st, 2nd, 4th, 5th, 7th, and 8th time-series images
        \item 4 images : 1st, 3rd, 5th, and 7th time-series images
        \item 3 images : 1st, 4th, and 8th time-series images
    \end{itemize}

Second, we selected images towards the end of the time series:
    \begin{itemize}
        \item 6 images : 3rd, 4th, 5th, 6th, 7th, and 8th time-series images
        \item 4 images : 5th, 6th, 7th, and 8th time-series images
        \item 3 images : 6th, 7th, and 8th time-series images
    \end{itemize}

This analysis helped us identify the minimum dataset size required to achieve reliable predictions, enabling us to optimize our data collection and processing efforts for future studies.

\subsubsection*{Top-2 Accuracy Assessment}
We also report Top-2 accuracy. This is because, in the context of soybean field classification based on field images, detailed classifications such as 7 classes can often be ambiguous due to the variability of environmental conditions and the vague boundaries between classes. Top-2 accuracy was calculated by counting incorrect classifications that were at most one class away from true as correct. For example: if our model predicted a plot as belonging to class 3 or 5 but was truly a class 4 plot, both of these predictions would be considered "correct". This was done to assess how well the prediction model was able to handle cases where a plot may be close to the true class but was misclassified. This data offers a practical analysis for soybean breeders and gives them confidence in the model's ability to not misclassify plots by a wide margin, i.e., more than one class away from true.

\section{Results and Discussion}

\subsection{Data description and exploration of seed yield and greenness loss}

\begin{figure}[H]
    \centering
    \begin{subfigure}[b]{0.49\linewidth}
        \centering
        \includegraphics[width=\linewidth]{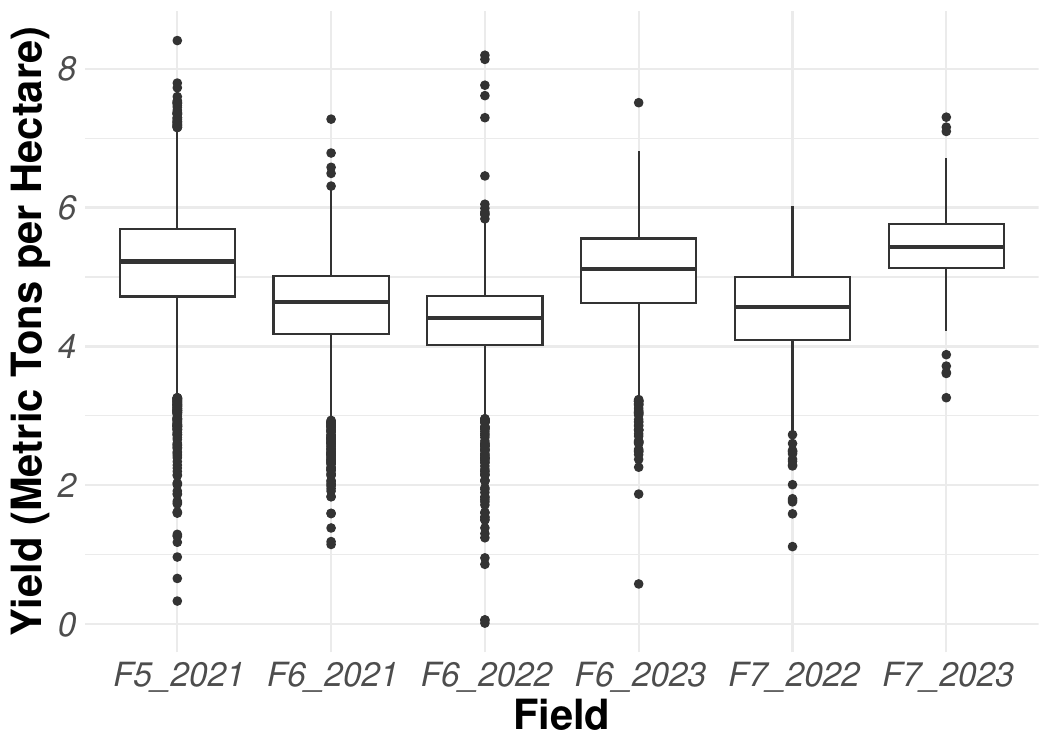}
        \caption{Distribution of yield for all plots across the six fields collected with UAS.}
        \label{fig:yield_distribution}
    \end{subfigure}
    \hfill
    \begin{subfigure}[b]{0.49\linewidth}
        \centering
        \includegraphics[width=\linewidth]{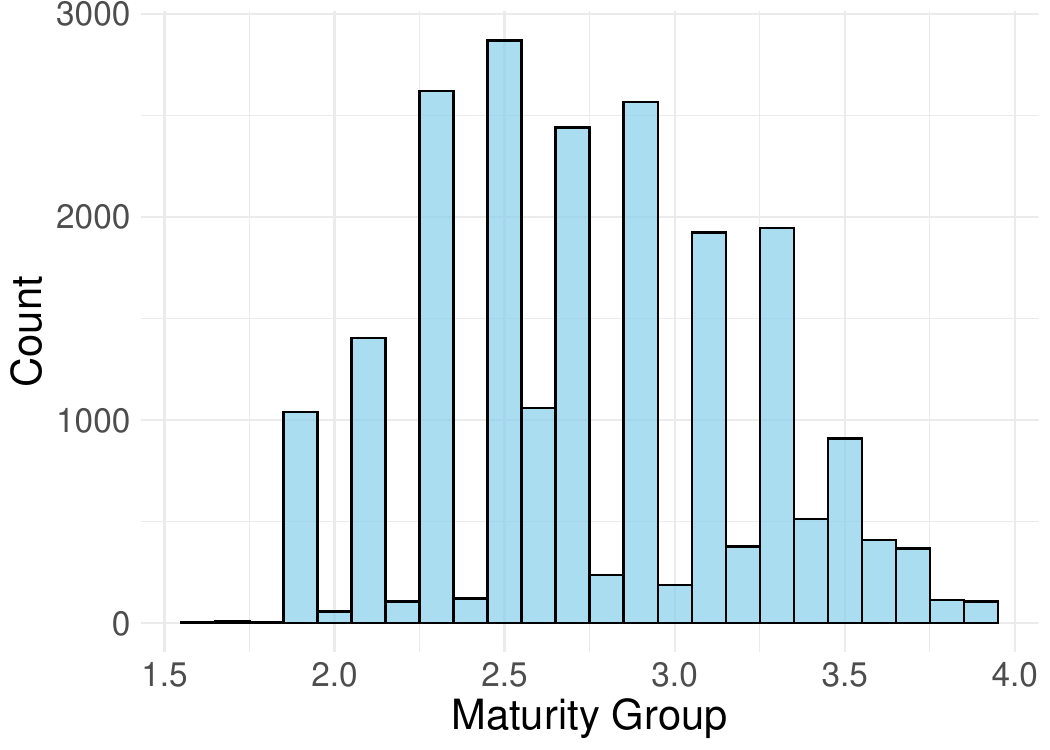}
        \caption{Frequency of RM groups collected for all plots across the six fields collected with UAS.}
        \label{fig:mg_histogram}
    \end{subfigure}
\caption{Distribution of ground truth RM scores for all six fields used for maturity classification training and testing (a), and distribution of yields for these fields (b). }
    \label{fig:RM_yield_distribution}    
\end{figure}

Our data consisted of ground truth RM ratings of between 1.6 - 3.9. The yield for each field averaged 5.15 MTH (Metric tonnes per hectare) with a range of 0.33-8.41 MTH for the 2021 F$_5$ field, 4.55 MTH with a range of 1.14-7.27 MTH for the 2021 F$_6$ field, 4.34 MTH with a range of 0.01-8.19 MTH for the 2022 F$_6$ field, 4.50 MTH with a range of 1.12-6.03 MTH for the 2022 F$_7$ field, 5.05 MTH with a range of 0.58-7.51 MTH for the 2023 F$_6$ field, and 5.43 MTH with a range of 3.26-7.30 MTH for the 2023 F$_7$ field. Figure \ref{fig:RM_yield_distribution} presents the distribution of RM ratings and yields for all six fields. 

\begin{figure}[htbp] 
    \centering
    \begin{subfigure}[b]{0.45\linewidth}
        \centering
        \includegraphics[width=\linewidth]{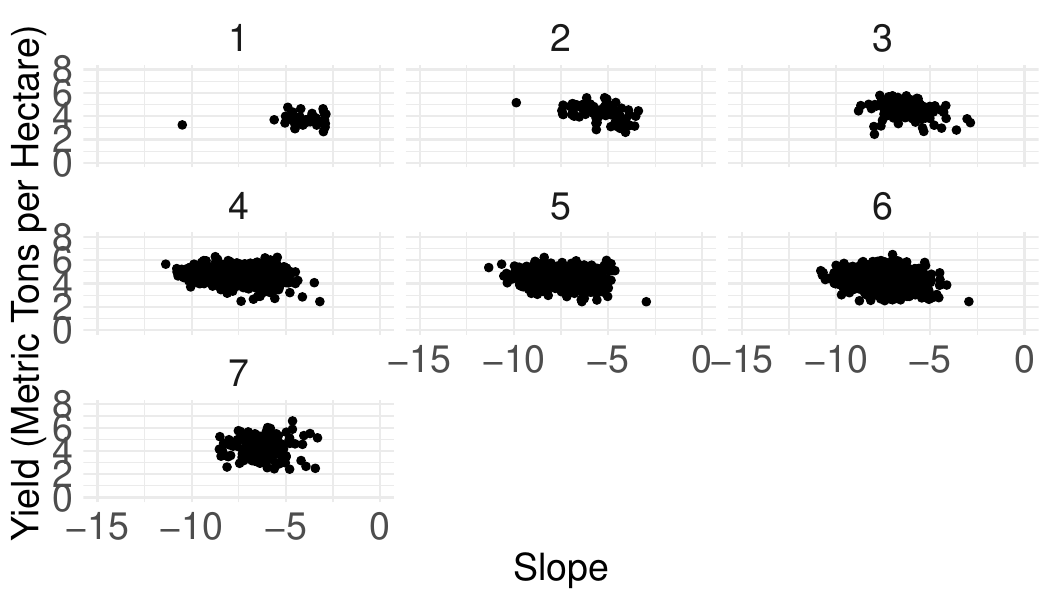}
        \caption{2021, F$_6$ material}
        \label{fig:cor_coe_b21}
    \end{subfigure}
    \hfill
    \begin{subfigure}[b]{0.45\linewidth}
        \centering
        \includegraphics[width=\linewidth]{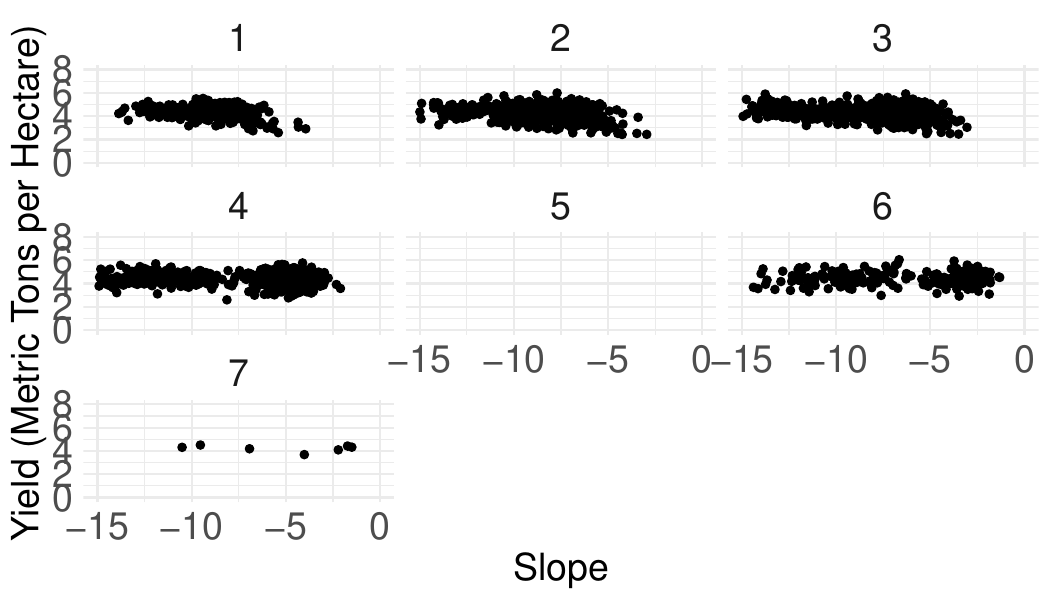}
        \caption{2022, F$_6$ material}
        \label{fig:cor_coe_b22}
    \end{subfigure}
    
    \vspace{0.5cm} 
    
    \begin{subfigure}[b]{0.45\linewidth}
        \centering
        \includegraphics[width=\linewidth]{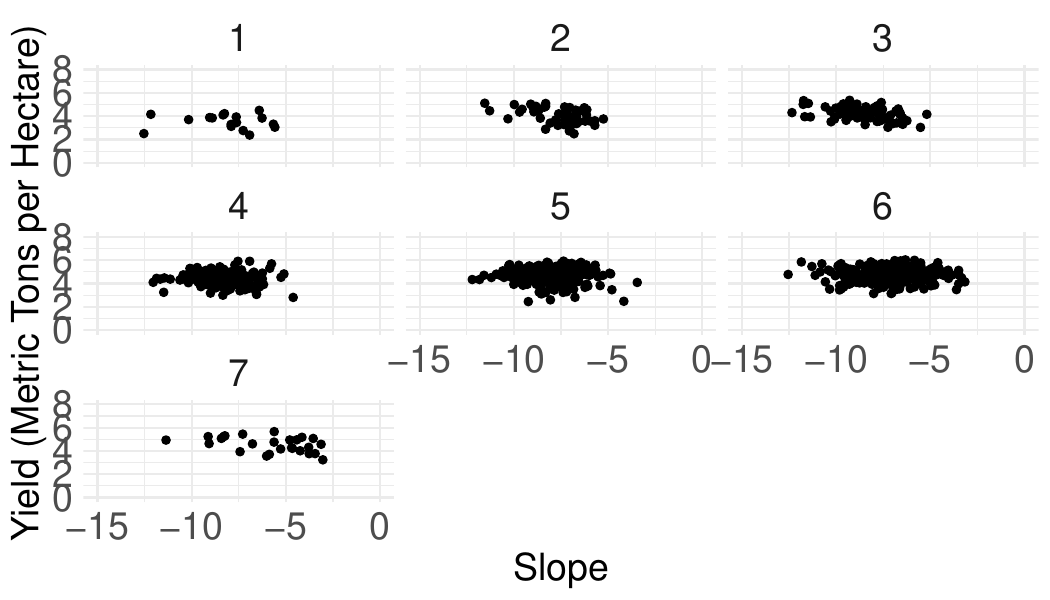}
        \caption{2022, F$_7$ material}
        \label{fig:cor_coe_c22}
    \end{subfigure}
    \hfill
    \begin{subfigure}[b]{0.45\linewidth}
        \centering
        \includegraphics[width=\linewidth]{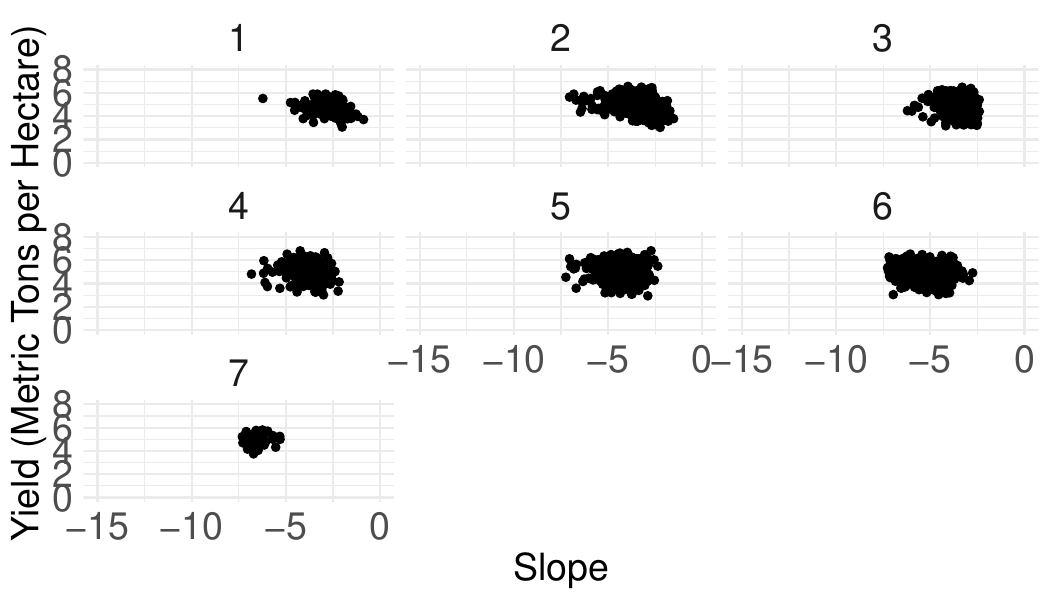}
        \caption{2023, F$_6$ material}
        \label{fig:cor_coe_b23}
    \end{subfigure}
    
    \vspace{0.5cm}
    
    \begin{subfigure}[b]{0.45\linewidth}
        \centering
        \includegraphics[width=\linewidth]{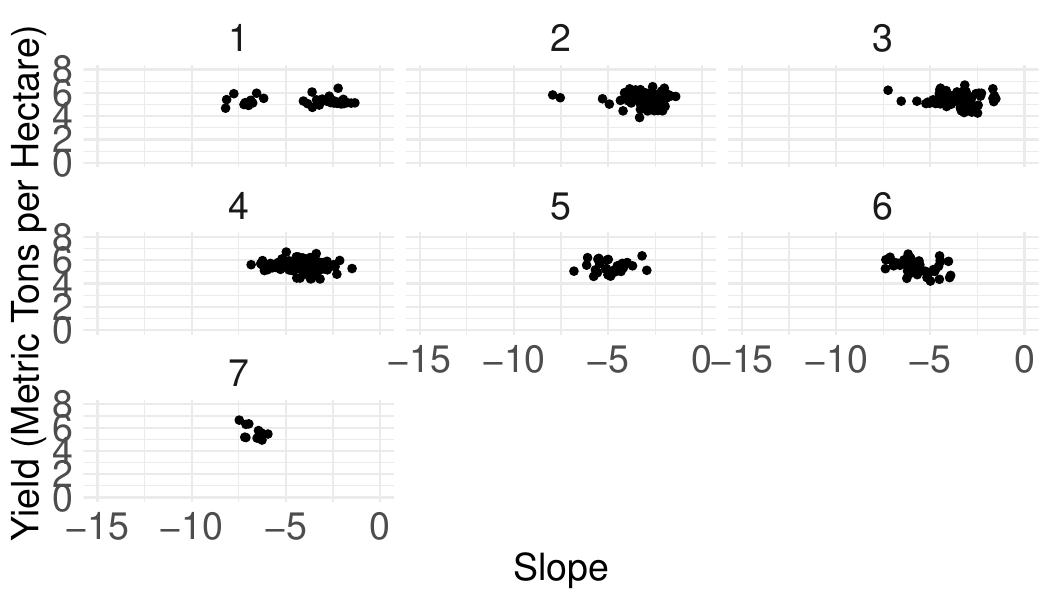}
        \caption{2023, F$_7$ material}
        \label{fig:cor_coe_c23}
    \end{subfigure}
    
    \caption{Correlation between slope and yield (tons/hectare) within RM groups for F$_6$ material from 2021 (a), F$_6$ material from 2022 (b), F$_7$ material from 2022 (c), F$_6$ material from 2023 (d), and F$_7$ material from 2023 (e). Data has been filtered to exclude values more than three standard deviations from the mean for both yield and slope.}
    \label{fig:combined_cor_coe}
\end{figure}

Before investigating data-driven approach to predict relative maturity, we examined the correlation between greenness loss slope (represented by the slope of decline in ExG value over time) and seed yield within each RM group for each field [Figure \ref{fig:combined_cor_coe}]. This allowed us to compare varieties with similar maturity dates to determine the relationship between greenness loss slope and seed yield. Within most RM groups, there is a negative correlation between greenness loss slope value and seed yield, i.e., a steeper greenness loss slope tends to exhibit higher yields (Table \ref{tab:cor_coe_sig}). This trend is more pronounced in earlier RM groups and reflects that varieties with a lower slope (negative value), i.e., rapid greenness loss after a prolonged greenness, are higher yielding (Figure \ref{fig:combined_cor_coe}). Observations of delayed leaf senesce and prolonged photosynthetic activity have been associated with higher yields in soybean \citep{wang2024comparative}.

One potential explanation for this relationship is the presence of stay-green traits in soybean. Of the previously defined types of stay-green \citep{thomas2000five}, Type A describes a plant that delays the loss of chlorophyll content and maintains photosynthetic activity longer than a typical non-stay-green plant. The expectation is that, given two soybean varieties that reach maturity on the same date, the variety with a steeper greenness loss slope may exhibit some of these stay-green traits and may have a higher yield since it was able to maintain photosynthetic activity for longer. This stay-green trait may explain some of the relationship between a steeper greenness loss slope and higher seed yields. Positive relationships between stay-green traits and yield have been shown in sorghum \citep{jordan2012relationship}, wheat \citep{christopher2008developmental}, and rice \citep{ba2009stay}. However, a previous study in soybeans produced mixed results, finding either no association or even a negative association between stay-green traits and seed yield \citep{luquez2001effects}. Our study included $> 22,000$ unique plot data, allowing us to investigate trait relationships with a higher sample size. As we previously noted, our study was conducted in central IA, U.S.A., across three years, so it is possible that the trends we observe in maturity categories 2 (RM 2.1-2.3), 3 (RM 2.4-2.6), and 4 (RM 2.7-2.9) were due to lack of a rapid temperature decline in the R7 growth stage, while later RM groups (3.0-3.2, 3.3-3.5 and 3.6-3.9) may have a forced senescence. By selecting varieties with a steeper greenness loss slope, breeders may improve yield potential by identifying varieties that initiate senescence later, thereby maintaining productivity for longer while reaching full maturity simultaneously with earlier senescing varieties \citep{fleitas2023functional}. Overall, the correlation between a steeper greenness loss slope and seed yield serves as an additional selection criterion for soybean breeders to make selections \citep{kamal2019stay, wang2021chlorophyll}. However, there is a need for further research in this area across varying latitudes for crops that are photo-period sensitive. 

These findings necessitate the need to investigate the relationship between greenness decline rate and relative maturity, and feasibility of data-driven approaches using UAS images to directly predict relative maturity in breeding and production applications.  

\begin{table}[H]
    \centering
    \caption{Correlation coefficients between ExG slope and yield within RM groups for each field. ** indicates a P-value $\leq$ 0.01 and * indicates a P-value of $\leq$ 0.05.}
    \begin{tabular}{cccccc}
    \hline
    \textbf{RM Group} & \textbf{2021 F$_6$} & \textbf{2022 F$_6$} & \textbf{2022 F$_7$} & \textbf{2023 F$_6$} & \textbf{2023 F$_7$} \\
    \hline
    1 & 0.08    & -0.32**   & -0.04     &-0.25**  & -0.01 \\
    2 &-0.40**  & -0.23**   & -0.44**   &-0.25**  & -0.11 \\
    3 & -0.29** & -0.21**   &-0.38**    &-0.12**  & -0.06 \\
    4 & -0.21** &-0.10**    & -0.12     &-0.16**  & -0.07 \\
    5 & -0.15** & -        & -0.05     &-0.04  &0.08 \\
    6 & -0.14** & 0.08      &-0.01      &-0.12*  &-0.38**  \\
    7 & 0.00    & -0.11     &  -0.43*   &0.18   & -0.52 \\
    \hline
    \end{tabular}
    \label{tab:cor_coe_sig}
\end{table}

\subsection{Relationship Between Greenness Decline Rate and Relative Maturity}
We examined the relationship between the rate of greenness loss, represented by the slope of decline in ExG value over time, and the RM group. This analysis was performed on the F$_6$ filial generation in 2021, F$_6$ and F$_7$ material in 2022, and F$_6$ and F$_7$ material in 2023. 

\begin{figure}[H] 
    \centering
    \begin{subfigure}[b]{0.45\linewidth}
        \centering
        \includegraphics[width=\linewidth]{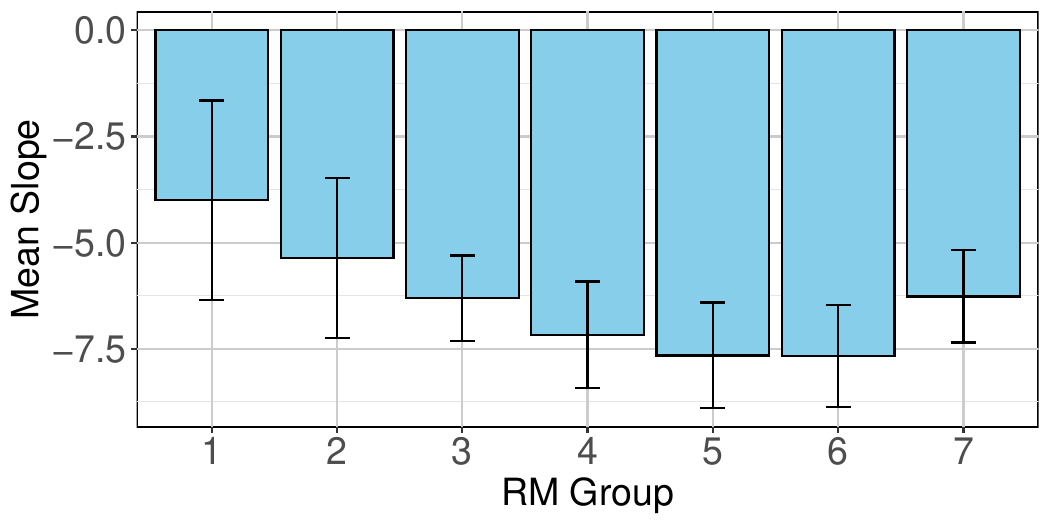}
        \caption{2021, F$_6$ material}
        \label{fig:slope_results_b21}
    \end{subfigure}
    \hfill
    \begin{subfigure}[b]{0.45\linewidth}
        \centering
        \includegraphics[width=\linewidth]{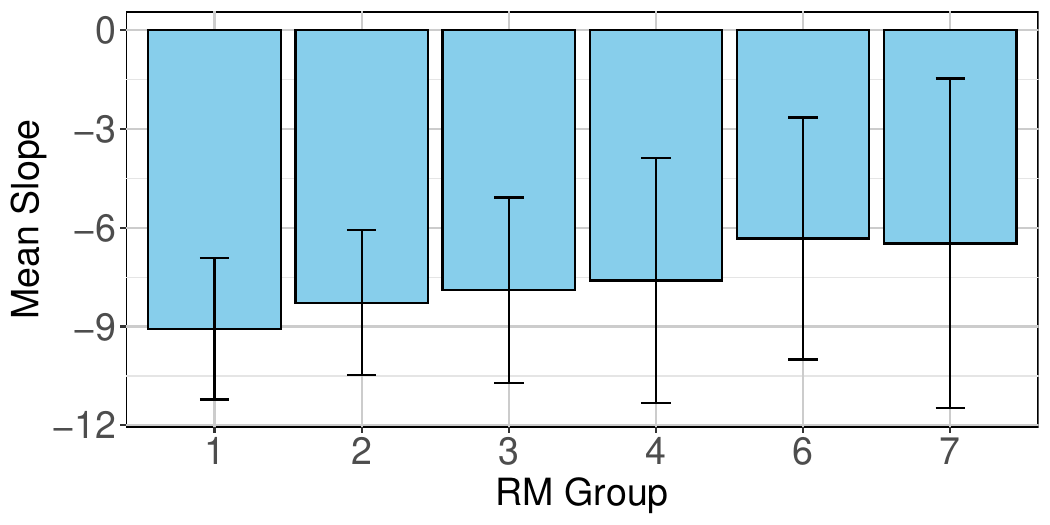}
        \caption{2022, F$_6$ material}
        \label{fig:slope_results_b22}
    \end{subfigure}
    
    \vspace{0.5cm}
    
    \begin{subfigure}[b]{0.45\linewidth}
        \centering
        \includegraphics[width=\linewidth]{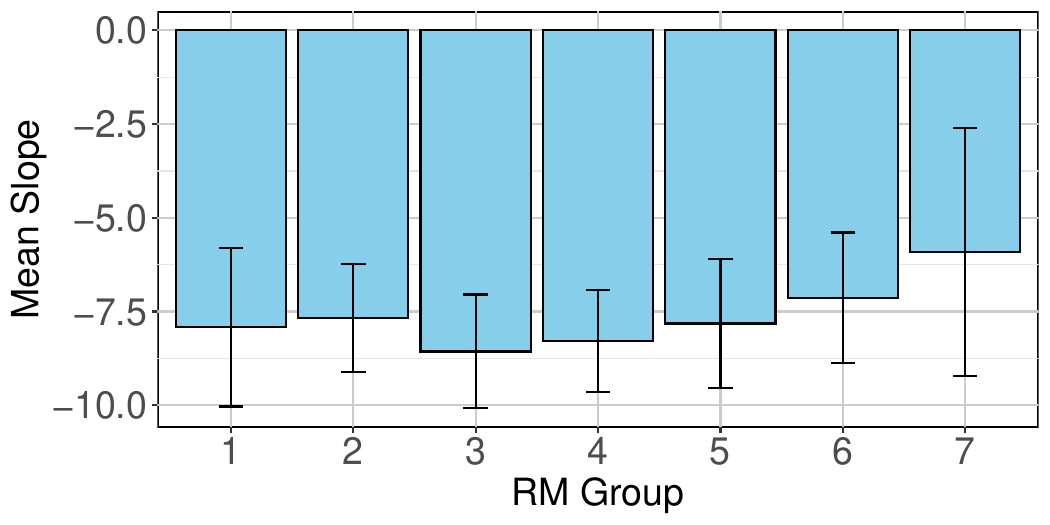}
        \caption{2022, F$_7$ material}
        \label{fig:slope_results_c22}
    \end{subfigure}
    \hfill
    \begin{subfigure}[b]{0.45\linewidth}
        \centering
        \includegraphics[width=\linewidth]{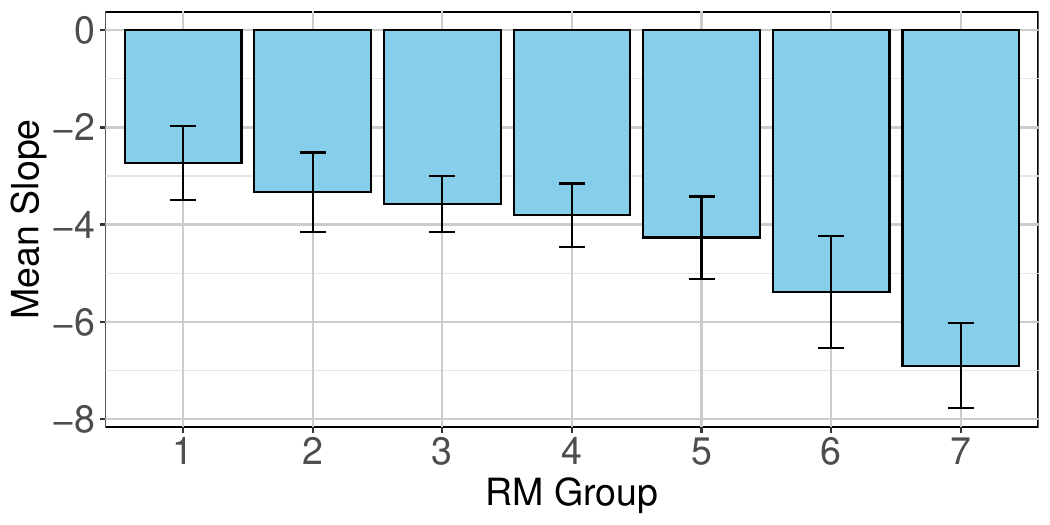}
        \caption{2023, F$_6$ material}
        \label{fig:slope_results_b23}
    \end{subfigure}
    
    \vspace{0.5cm}
    
    \begin{subfigure}[b]{0.45\linewidth}
        \centering
        \includegraphics[width=\linewidth]{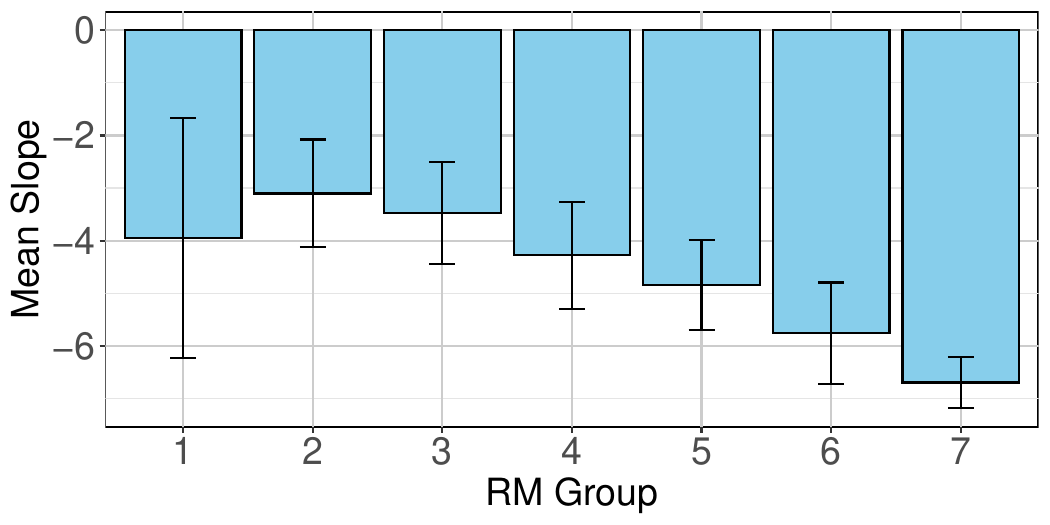}
        \caption{2023, F$_7$ material} 
        \label{fig:slope_results_c23}
    \end{subfigure}
    
    \caption{Histograms reporting mean slope values for different RM groups with standard deviation error bars for F$_6$ material from 2021 (a), F$_6$ material from 2022 (b), F$_7$ material from 2022 (c), F$_6$ material from 2023 (d), and F$_7$ material from 2023 (e).}
    \label{fig:combined_slope_histograms}
\end{figure}

\begin{figure}[H] 
    \centering
    \begin{subfigure}[b]{0.45\linewidth}
        \centering
        \includegraphics[width=\linewidth]{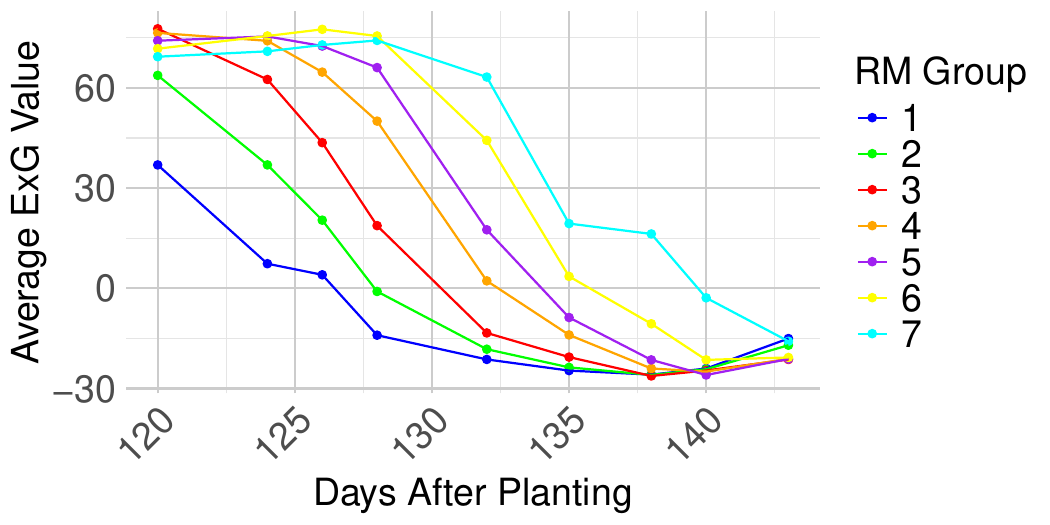}
        \caption{2021, F$_6$ material}
        \label{fig:ave_exg_b21}
    \end{subfigure}
    \hfill
    \begin{subfigure}[b]{0.45\linewidth}
        \centering
        \includegraphics[width=\linewidth]{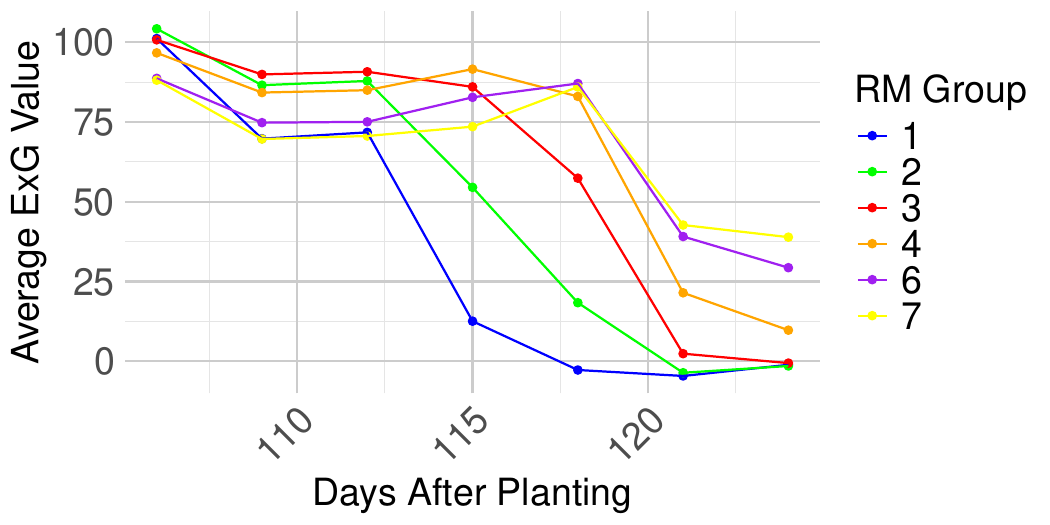}
        \caption{2022, F$_6$ material}
        \label{fig:ave_exg_b22}
    \end{subfigure}
    
    \vspace{0.5cm} 
    
    \begin{subfigure}[b]{0.45\linewidth}
        \centering
        \includegraphics[width=\linewidth]{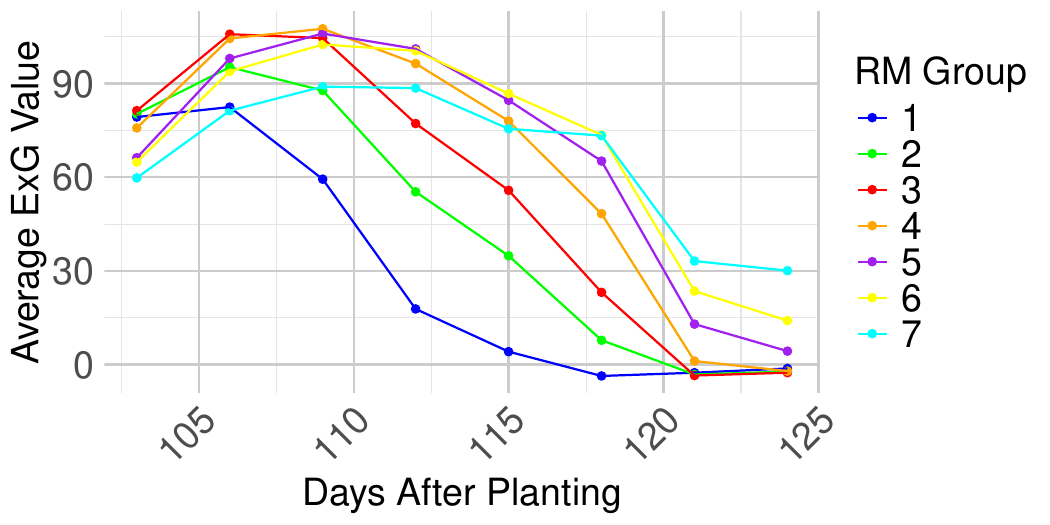}
        \caption{2022, F$_7$ material}
        \label{fig:ave_exg_c22}
    \end{subfigure}
    \hfill
    \begin{subfigure}[b]{0.45\linewidth}
        \centering
        \includegraphics[width=\linewidth]{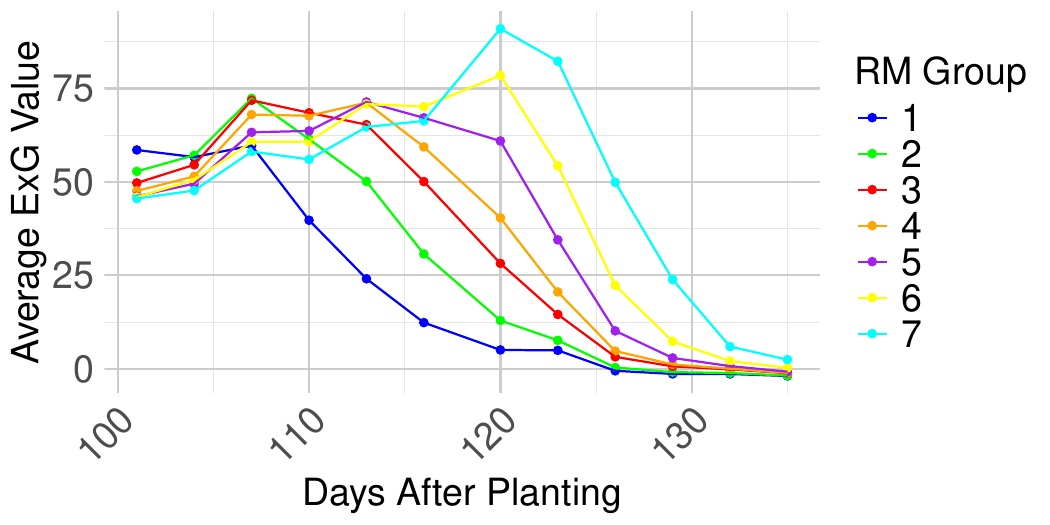}
        \caption{2023, F$_6$ material}
        \label{fig:ave_exg_b23}
    \end{subfigure}
    
    \vspace{0.5cm} 
    
    \begin{subfigure}[b]{0.45\linewidth}
        \centering
        \includegraphics[width=\linewidth]{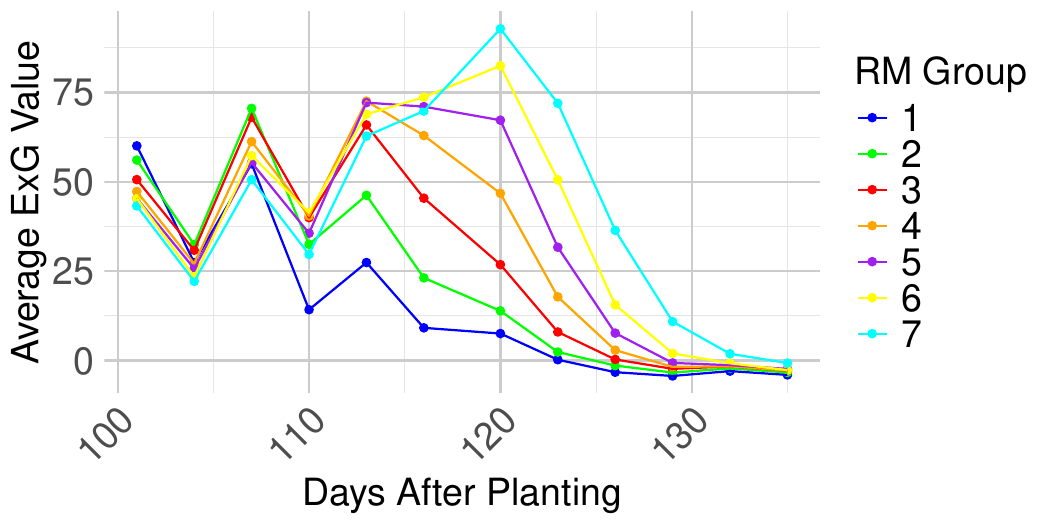}
        \caption{2023, F$_7$ material}
        \label{fig:ave_exg_c23}
    \end{subfigure}
    
    \caption{Average ExG value for seven RM groups over time for F$_6$ material from 2021 (a), F$_6$ material from 2022 (b), F$_7$ material from 2022 (c), F$_6$ material from 2023 (d), and F$_7$ material from 2023 (e). Note the differences between the 2021 and 2023 data compared to 2022: in 2021 and 2023, the average ExG values for all RM groups converge to a similar point, whereas the 2022 RM group values do not.}
    \label{fig:combined_ave_exg}
\end{figure}

Analysis of the greenness loss rate in 2021 and 2023, as indicated by the slope, reveals a consistent trend across different RM classes Figure \ref{fig:combined_slope_histograms}. These results suggest that soybean varieties in later RM groups exhibit a smaller, steeper greenness loss slope than those belonging to earlier RM groups, i.e., varieties with later maturity dates tend to lose greenness more rapidly than those with earlier maturity dates. However, this trend is not observed in the field from 2022 F$_7$ where a minimal relationship exists between slope and RM class. The average ExG line graph for the F$_7$ material in 2022 shows that the later RM classes were still undergoing greenness loss and had not fully converged with the early RM varieties [Figure \ref{fig:ave_exg_c22}]. This contrasts with the data from 2021 and 2023, where all RM classes appeared to reach a similar final ExG value [Figure \ref{fig:combined_ave_exg}]. In the 2022 F$_6$ field, a strong but opposite relationship was observed, where early RM lines tended to have a steeper greenness loss slope. These results emphasize the utility of integrating rate of color change (i.e. hue changes) with sophisticated machine learning models to make predictions of relative maturity.

\subsection{Model Performance of Deep Learning Architectures for Maturity Classification}
For an automated soybean maturity classification, we evaluated several pre-trained neural network models, including ResNet18, ResNet34, ResNet50, VGG16, and MobileNet, across different configurations. Each model was subjected to adjustments in hyperparameters and trained with datasets from different years (2021, 2022, and 2023) as well as a combined time-series dataset [8 time points]. Moreover, models were trained to classify maturities in four [1: 1.6-2.0, 2: 2.1-2.3, 3: 2.4-3.2, 4: 3.3-3.9], five [0: 1.6-2.0, 1: 2.1-2.3, 2: 2.4-2.6, 3: 2.7-3.2, 4: 3.3-3.9], and seven [0: 1.6-2.0, 1: 2.1-2.3, 2: 2.4-2.6, 3: 2.7-2.9, 4: 3.0-3.2, 5: 3.3-3.5, 6: 3.6-3.7, 7: 3.8-3.9] classes. This was done for practical application in a breeding program, where similar maturities are grouped together in the creation of a test to ensure like-to-like comparison based on relative maturity among varieties.

ResNet34 demonstrated slightly better performance than other neural network models, particularly when trained with the combined dataset and across multiple class configurations. Specifically, in the case of 5-class classification, ResNet34 achieved an accuracy of 78.05\%, outperforming GoogleNet (76.61\%), DenseNet (76.96\%), and MobileNet (76.82\%). This suggests that ResNet34 effectively captures the nuances of soybean growth variations across multiple seasons, benefiting from a broader training dataset that enhances its generalization capabilities. The variation in the number of classes allowed us to further explore the trade-off between model complexity and classification accuracy, with more classes providing finer granularity at the cost of increased model complexity.

\subsection{Performance on Different Classification Labels}
\begin{figure}[htbp]
  \centering
  \begin{subfigure}[b]{0.30\textwidth}
    \includegraphics[width=\textwidth]{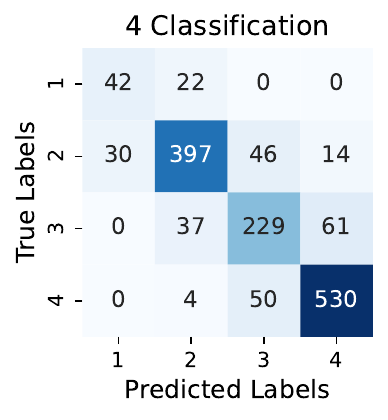}
    \caption{}
    \label{fig:image1}
  \end{subfigure}
  \hfill
  \begin{subfigure}[b]{0.30\textwidth}
    \includegraphics[width=\textwidth]{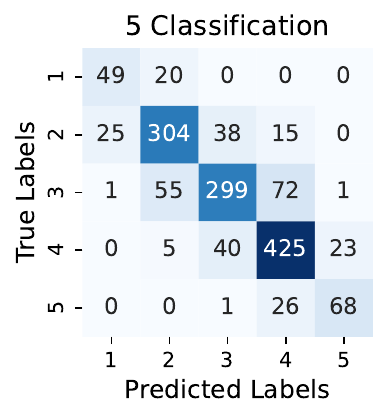}
    \caption{}
    \label{fig:image2}
  \end{subfigure}
  \hfill
  \begin{subfigure}[b]{0.38\textwidth}
    \includegraphics[width=\textwidth]{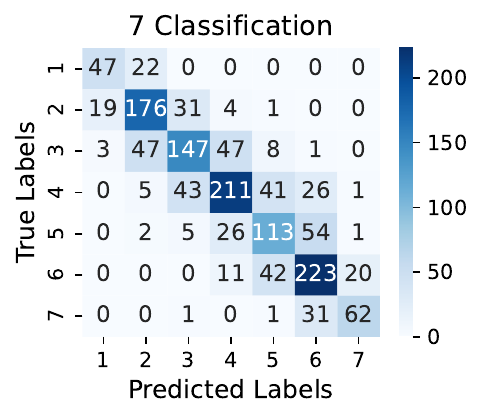}
    \caption{}
    \label{fig:image3}
  \end{subfigure}
  \caption{This figure presents the confusion matrices for the ResNet34 model applied to soybean maturity classification across different class configurations: (a) 4-class, (b) 5-class, and (c) 7-class.}
  \label{fig:result_confusionmatrix}
\end{figure}

For the 4-classification, the model demonstrated its capability to effectively differentiate between fewer maturity stages with significant accuracy as shown in Figure \ref{fig:result_confusionmatrix}. When utilizing data from the years 2021 and 2022, the model achieved its highest accuracy of 84\% (Table \ref{tab:totalresult}), indicating robust performance under the first set of criteria. Additionally, under a second set of criteria, which included data from 2021, 2022, and the extended datasets of 2023, the model maintained an accuracy of 79\%. For the 5-class classification, using the combined datasets from 2021, 2022, and 2023, the model achieved an accuracy of 78\%. When the same datasets were used for a more detailed 7-class classification, the accuracy was 66.5\%. Therefore, for the 7-class classification, we implemented hierarchical classification to enhance the model's ability to handle more complex and detailed class structures. The hierarchical classification approach proved effective, improving the model's performance. 
Specifically, in the 7-classification case, the hierarchical classification method led to an accuracy improvement of approximately 6\% compared to directly classifying into 7 categories without using a hierarchical approach. This demonstrates that the hierarchical classification effectively simplifies the complexity of detailed class divisions, making the model more adaptable to challenging classification tasks.
However, while the model shows the capability to manage more complex class divisions, the increase in classification detail can negatively impact overall accuracy due to the added complexity. To further enhance performance, the availability of larger datasets would be beneficial for training a more robust and generalizable model.

In addition to the standard accuracy metrics, top-2 accuracy was evaluated to provide further insight into the model's predictive capabilities. For the 4-class model in Table \ref{tab:totalresult}, the top-2 accuracy reached 99.1\%, indicating a high likelihood of the true class being within the two most probable predictions made by the model. Similarly, the 5-class and 7-class models demonstrated top-2 accuracies of 98.7\% and 96.2\%, respectively. This metric is particularly relevant in the context of detailed classifications, where the distinction between adjacent maturity stages can be subtle, yet the model reliably identifies the correct stage within its top two predictions. It prevents breeding programs from making errors in misclassifying relative maturity and entering them in incorrect tests the following season.

\begin{table}[h]
\centering
\caption{Accuracy results based on different classification-dataset schemes.}
\label{tab:totalresult}
\begin{tabular}{@{}lcrrr@{}} 
\toprule[1.0pt] 
Classification & Dataset & Train Accuracy & Test Accuracy & Top-2-Accuracy \\
\midrule[0.75pt] 
{4 Classes(1st)} & 2021, 2022 & 96.84 & 84.92 & 99.60 \\
{4 Classes(1st)} & 2021, 2022, 2023 & 95.71 & 81.94 & 99.10 \\
{4 Classes(2nd)} & 2021, 2022, 2023 & 90.56 & 79.09 & 99.10 \\
{5 Classes} & 2021, 2022, 2023 & 96.32 & 78.05 & 98.70 \\
{7 Classes} & 2021, 2022, 2023 & 82.91 & 66.50 & 96.22 \\
{7 Classes - Hierarchical} & 2021, 2022, 2023 & 88.52 & 72.98 & 97.22 \\
\bottomrule[1.0pt] 
\end{tabular}
\end{table}

\subsection{Performance on Various Subsets of Temporal Imagery}
\begin{figure}[htbp]
  \centering
  \begin{subfigure}[b]{0.49\textwidth}
    \includegraphics[width=\textwidth]{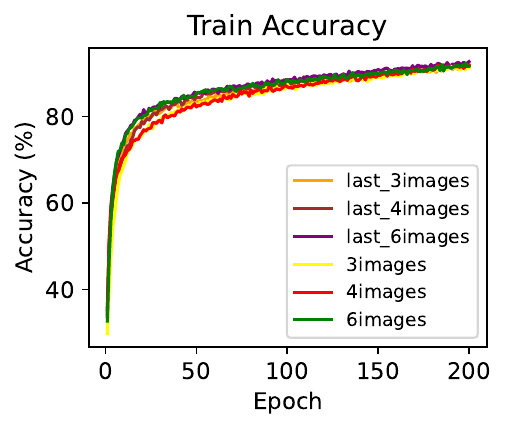}
    \caption{Train accuracy on different subsets}
    \label{fig:result_diff_timepoint}
  \end{subfigure}
  \hfill
  \begin{subfigure}[b]{0.49\textwidth}
    \includegraphics[width=\textwidth]{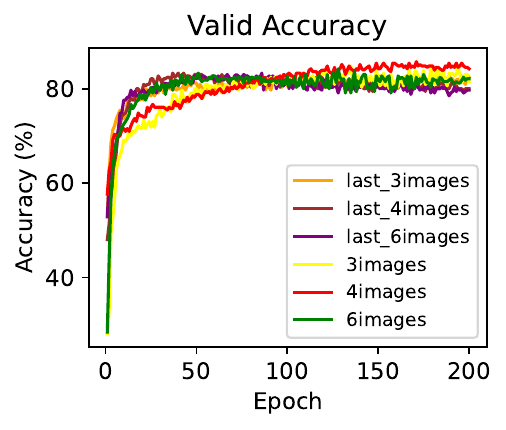}
    \caption{Test accuracy on different subsets}
    \label{fig:image2}
  \end{subfigure}
  \caption{(a) displays the training accuracy over epochs, showing a consistent trend across different dataset sizes. Despite the variations in the number of datasets used, the general trajectory of training performance remains similar. (b) illustrates that model performance is comparably stable across different subsets of the time series dataset.}
  \label{fig:two_images}
\end{figure}

Exploring the necessity and sufficiency of temporal data, we conducted experiments to identify the minimal set of time-point images required to maintain high classification accuracy. By reducing the dataset from eight images to subsets of six, four, and three images per plot, we observed that accuracy for 4 classifications (4 classes - 1st) remained robust across these subsets (Figure \ref{fig:result_diff_timepoint} and Table \ref{tab:result_diff_timepoint}). That is, by reducing drone flights from eight to just three, we can still achieve nearly identical classification performance, effectively halving the number of necessary flights for maturity classification. 

For example, by using subsets of 6, 4, or 3 images from the original set of eight time-series images, we consistently achieved a comparable accuracy of approximately 84\%, indicating that the full 8 time-series set is not required to maintain similar levels of training and testing performance. Notably, using the last 6 time-series images resulted in the highest performance with an accuracy of 88\%, even surpassing the accuracy achieved with all 8 images.
At the same time, since robust performance can be achieved with three time-series data, the minimum number of time-series data required for reliable performance can be estimated.
This suggests that reduced image subsets can be effectively used without compromising model performance, enabling breeders to estimate the minimum number of drone flights required for drone-based imagery. Breeders can reduce the number of necessary flights from eight to as few as three, significantly enhancing operational efficiency and resource utilization. This result has significant implications for operational efficiencies in a breeding program utilizing drone-based imaging for maturity classification, offering cost reductions and logistical simplifications without compromising the accuracy of maturity assessments for test creation and variety placement.

\begin{table}[h]
    \centering
    \caption{Performance on Different Number of Time-point}
    \label{tab:result_diff_timepoint}
    \begin{tabular}{ccc} 
        \toprule
        \textbf{Num. of Time-point} & \textbf{Train Acc} & \textbf{Test Acc} \\ 
        \midrule
        8 images & 96.80 & 84.92\\
        6 images & 91.60 & 84.22\\
        4 images & 91.76 & 83.92\\
        3 images & 91.29 & 84.52\\
        6 last images & 92.67 & 88.39\\
        4 last images & 91.80 & 84.82\\
        3 last images & 92.00 & 84.82\\
        \bottomrule
    \end{tabular}
\end{table}

\section{Conclusion}
We present a method for remotely estimating soybean relative maturity (RM) using UAS and machine learning. We achieve this using a novel method of tracking plot greenness over time with the creation of a 2-dimensional contour plot, allowing for a more efficient data type for model training. We achieved model accuracy of up to 85\% in predicting soybean RM. This number jumps up to 99\% when we allow for misclassification of one away from true. We also show that as few as three time points can be used without significantly decreasing prediction accuracy. Finally, we explored the rate of greenness loss as represented by the loss of ExG value over time and its relationship to soybean RM and yield. Results show that later RM groups tend to have steeper, more negative greenness loss slopes, while yield tends to have a negative correlation to greenness loss slope. 

The implications of this work on soybean breeding programs are significant, as it can save labor hours, which can now be allocated elsewhere in the program. Our methodology also provides a unique approach to soybean maturity classification. While previous works tend to rely on simple binary "mature" versus "immature" classification schemes, we offer a more nuanced approach using up to seven classes. We also explore the ideal number of flights needed to achieve high prediction accuracies. Decreasing the number of flights from eight to three can further decrease the number of labor hours needed for maturity data collection. This time savings is compounded for certain soybean breeding programs that can have tens if not hundreds of thousands of plots spread across a large region. Coordinating dozens of raters to collect data on each of these plots is impractical. Sending a single drone pilot who can capture the same number of plots in a fraction of the time makes this data collection more practical and will not suffer from inter- or intra-rater error. 

While our work presents promising results for the mid-west region, further experiments should be done to test the robustness of our model on soybean materials belonging to different MG groups. Trends apparent in the 2D contour plots that are used in this experiment may be different (as expected) for MG soybean lines of much earlier or much later maturity, requiring new training and testing datasets from these soybean lines. Furthermore, the relationship observed between the greenness loss slope and RM group and greenness loss slope and yield will also require further experimentation on different MG group soybean lines to be confirmed. This is especially true for latitudes that are not likely to encounter a killing frost later in the growing season.

\section*{Acknowledgments}
The authors are thankful to Brian Scott, Jennifer Hicks, Ryan Dunn and David Zimmerman for their efforts in field experiments. We also thank the many graduate and undergraduate students who assisted in data collection. 

\subsection*{Conflicts of Interest}
The authors declare no conflicts of interest with this work.

\section{Supplementary Materials}

\subsection*{Author Contributions} 
B.K.: Methodology; Software; Formal analysis; Data Curation; Visualization; Writing - Original Draft; Writing - Review \& Editing; \\
S.W.B.: Conceptualization; Methodology; Software; Formal analysis; Investigation; Data Curation; Visualization; Writing - Original Draft; Writing - Review \& Editing; \\
T.Z.J.: Methodology; Software; Writing - Review \& Editing; \\
S.S.: Methodology; Funding acquisition; Writing - Review \& Editing; \\
A.S.: Methodology; Funding acquisition; Writing - Review \& Editing; \\
A.K.S.: Conceptualization; Methodology; Resources; Visualization; Supervision; Project administration; Funding acquisition; Writing - Original Draft; Writing - Review \& Editing; \\ 
B.G.: Conceptualization, Methodology; Resources; Visualization; Supervision; Project administration; Funding acquisition; Writing - Review \& Editing.

\subsection*{Funding}
The authors sincerely appreciate the funding support from Iowa Soybean Association (A.K.S.), USDA CRIS project IOW04714 (A.K.S., A.S.), AI Institute for Resilient Agriculture (USDA-NIFA \#2021-67021-35329) (B.G., S.S., A.S., A.K.S.), COALESCE: COntext Aware LEarning for Sustainable CybEr-Agricultural Systems (CPS Frontier \#1954556) (S.S., B.G., A.S., A.K.S.), USDA-NIFA FACT (\#2019-67021-29938) (A.S.), Smart Integrated Farm Network for Rural Agricultural Communities (SIRAC) (NSF S\&CC \#1952045) (A.K.S., S.S.), R. F. Baker Center for Plant Breeding (A.K.S., A.S.), Plant Sciences Institute (A.K.S., S.S., B.G.), and G.F. Sprague Chair in Agronomy.

\subsection*{Data Availability}
Data can be requested from A.K. Singh. 
Models and workflows are available on the \href{https://bitbucket.org/baskargroup/soystcontour/src/main/}{Git Repository}.

\bibliographystyle{apalike}
\bibliography{output}

\end{document}